\documentclass[%
 aip,
 amsmath,amssymb,
 reprint,%
]{revtex4-1}

\usepackage{graphicx}
\usepackage{dcolumn}
\usepackage{bm}

\usepackage[utf8]{inputenc}
\usepackage[T1]{fontenc}
\usepackage{mathptmx}
\usepackage{etoolbox}
\usepackage{svg}
\usepackage{verbatim} 
\usepackage{soul} 

\usepackage{enumitem} 
\usepackage{xcolor}
\definecolor{BrickRed}{rgb}{0.8,0.25,0.33}
\colorlet{BRICKRED}{BrickRed}
\usepackage[hidelinks]{hyperref}
\usepackage[capitalise]{cleveref} 
\usepackage{caption} 
\usepackage{subcaption} 
\usepackage{ragged2e} 
\usepackage{titlesec}
\titleformat{\section}
{\color{BrickRed}\normalfont\small\bfseries\MakeUppercase}
{\color{BrickRed}\thesection}{0pt}{}

\titleformat{\subsection}
{\normalfont\small\bfseries}
{\thesubsection}{0pt}{}

\titleformat{\subsubsection}
{\normalfont\small\itshape}
{\thesubsubsection}{0pt}{}

\titlespacing{\section}{0pt}{3ex plus .1ex minus .2ex}{0pt}
\titlespacing{\subsection}{0pt}{2ex plus .1ex minus .2ex}{0pt}
\titlespacing{\subsubsection}{0pt}{1.5ex plus .1ex minus .2ex}{0pt}

\titleformat{\SIsection}
{\color{BrickRed}\normalfont\small\bfseries\MakeUppercase}
{\color{BrickRed}\theSIsection}{0pt}{0pt}
\titlespacing{\SIsection}{0pt}{3ex plus .1ex minus .2ex}{0pt}

\newcommand{\supplementarysection}{%
  \setcounter{figure}{0} 
  \let\oldthefigure\thefigure 
  \renewcommand{\thefigure}{S\oldthefigure} 
  
  \section{Supplementary Material}
  \label{sec:Supplemental}

  \setcounter{secnumdepth}{3}
  \setcounter{section}{0}
  \setcounter{subsection}{0}
  \setcounter{subsubsection}{0}

  \newcommand{\thesubsection}{S\arabic{subsection}}
  \newcommand{\thesubsubsection}{\thesubsection.\arabic{subsubsection}}
}

\begin{document}

\title{Tailored Forecasting from Short Time Series via Meta-learning}


\author{Declan A. Norton}
\email[Correspondence email address: ]{nortonde@umd.edu}
\affiliation{Department of Physics, University of Maryland, College Park, Maryland 20742, USA\looseness=-1}

\author{Edward Ott}
\affiliation{Department of Physics, University of Maryland, College Park, Maryland 20742, USA\looseness=-1}
\affiliation{Institute for Research in Electronics and Applied Physics, University of Maryland, College Park, Maryland 20742, USA\looseness=-1}
\affiliation{Department of Electrical and Computer Engineering, University of Maryland, College Park, Maryland 20742, USA\looseness=-1}

\author{Andrew Pomerance}
\affiliation{Potomac Research LLC, Alexandria, Virginia 22314, USA\looseness=-1}

\author{Brian Hunt}
\affiliation{Institute for Physical Science and Technology, University of Maryland, College Park, Maryland 20742, USA\looseness=-1}
\affiliation{Department of Mathematics, University of Maryland, College Park, Maryland 20742, USA\looseness=-1}

\author{Michelle Girvan}
\affiliation{Department of Physics, University of Maryland, College Park, Maryland 20742, USA\looseness=-1}
\affiliation{Institute for Research in Electronics and Applied Physics, University of Maryland, College Park, Maryland 20742, USA\looseness=-1}
\affiliation{Institute for Physical Science and Technology, University of Maryland, College Park, Maryland 20742, USA\looseness=-1}
\affiliation{Santa Fe Institute, 
Santa Fe, New Mexico 87501, USA\looseness=-1}

\begin{abstract}
Machine learning models can effectively forecast dynamical systems from time-series data, but they typically require large amounts of past data, making forecasting particularly challenging for systems with limited history. To overcome this, we introduce Meta-learning for Tailored Forecasting using Related Time Series (METAFORS), which generalizes knowledge across systems to enable forecasting in data-limited scenarios. By learning from a library of models trained on longer time series from potentially related systems, METAFORS builds and initializes a model tailored to short time-series data from the system of interest. Using a reservoir computing implementation and testing on simulated chaotic systems, we demonstrate that METAFORS can reliably predict both short-term dynamics and long-term statistics without requiring contextual labels. We see this even when test and related systems exhibit substantially different behaviors, highlighting METAFORS' strengths in data-limited scenarios.
\end{abstract}


\maketitle 
\section{Introduction}\label{sec:Intro}
\noindent
Forecasting of dynamical systems is crucial across fields such as weather\cite{Price2025_MLWeather} and climate\cite{Arcomano2022_AtmHybrid} science, neuroscience\cite{Wein2022_ForecastingBrainActivity,DeMatola2025_BrainStateForecasting}, epidemiology\cite{Ray2023_USCOVIDEnsembles}, and finance\cite{Sezer2020_FinanceDLReview}. However, many systems lack accurate knowledge-based (e.g., mathematical or physical) models, necessitating data-driven approaches\cite{Brunton2019_Book,Han2021_Deep_TS_Prediction_Review,Brunton2016_SINDy,Rudy2017_PDESindy} -- either in standalone configurations or to augment insufficient models in hybrid configurations\cite{Pathak_Hybrid,Arcomano2022_AtmHybrid}. Standard machine learning (ML) forecasters, while powerful, are often `data intensive,' requiring extensive training data to function effectively. They are also frequently ‘brittle,’ struggling to generalize across systems, even in cases where the systems' dynamics are not very different\cite{Wang2021_BridgingPhysics_and_Data_for_DS,Goring2024_OutOfDomain}. For example, public health officials responding to a novel virus outbreak may find that early data are insufficient for training a new model, and models trained on previous outbreaks fail to generalize.

This work addresses the challenge of forecasting when only a short time series -- insufficient to train a standalone model -- is available from the system of interest. We assume access to longer time series from other related systems (and/or simulations) and that the long and short signals are both vector time series -- sequences of measurement vectors over time. 
Our method, drawing on principles from multi-task learning and meta-learning, leverages knowledge from these related datasets to enable prediction or improve prediction accuracy while mitigating issues of data-intensity and brittleness.

While traditional multi-task learning aims to improve robustness by training a single model across multiple datasets, it can dilute system-specific information\cite{MTL_Review2021,Yang2020_Transfer_Learning}. In contrast, meta-learning focuses on quickly adapting to new tasks by generalizing from task-specific models\cite{Hospedales2022_MetaL_in_NN_Survey,Brazdil2022_MetaL_Book,Lemke2015_MetaL_Survey,Pan2010_TLSurvey}. Meta-learning has been applied in a range of settings, including hyperparameter optimization\cite{Feurer2015_BayesianOpt_w_MetaL}, algorithm selection and combination\cite{Lemke2010_MetaL_ForecastCombination,Talagala2023_MetaL_for_Forecasting}, and few-shot  
learning\cite{Finn17_MAML,Yao2019_HSML,Raghu2019_ANIL,Rusu2019_LEO,Wu2018_MeLA,Joshaghani2023_RetailMetaL,Canaday2021MARC,Kirchmeyer2022a_CoDA,Yin2021_LEADS,Wang2022_DyAd,Panahi2025_CriticalTransitions,Oreshkin2021_MetaL_Zeroshort_TSF}, which is our focus. Specifically, we wish to tailor the parameters of an ML model to forecast a dynamical system for which only a small amount of data is available.

For individual time series prediction tasks, we focus on ML models with memory.  By `memory,' we mean that the output of the ML model depends not only on measurements of the current state of the system being forecast, but also on measurements at previous times.  Such models include those with intrinsic memory encoded in an internal state -- e.g., those based on recurrent neural networks, including long short-term memory models (LSTMs), gated recurrent units (GRUs), and reservoir computers (RCs) -- and those without intrinsic memory that explicitly include time-delayed measurements in their input -- e.g., next-generation reservoir computers (NGRCs) and certain feedforward neural networks and kernel machines. While memory-based ML models are well suited for time series prediction, they face an additional challenge when forecasting from short time series: accurate predictions rely on proper initialization of the model's memory and often require substantial data from immediately before the start of the forecast\cite{Gauthier2021_NGRC,Zhang2023_Catch22,Lu2018_Attractor_Reconstruction,Grigoryeva2024_ColdStartRC,Kemeth2021_ColdStartLSTM}. We note that the data requirements for memory initialization and for forecaster training are different. It is possible to have enough data to initialize the model's forecast without having enough to train its parameters. It is also possible to have enough data from the system of interest to train a good model while lacking sufficient data immediately before the start of the forecast, e.g., when using the model to forecast from a new short times series in cases where the underlying dynamics are expected to be the same as those seen in training. Without sufficient data for initialization, memory-based models struggle to produce useful forecasts from a `cold start,' making forecasts inaccurate or unattainable. We refer to a suitable initialization of the model's memory as a `cold-start vector.'

\begin{figure*}
	\centering
	\includegraphics[width=\linewidth]{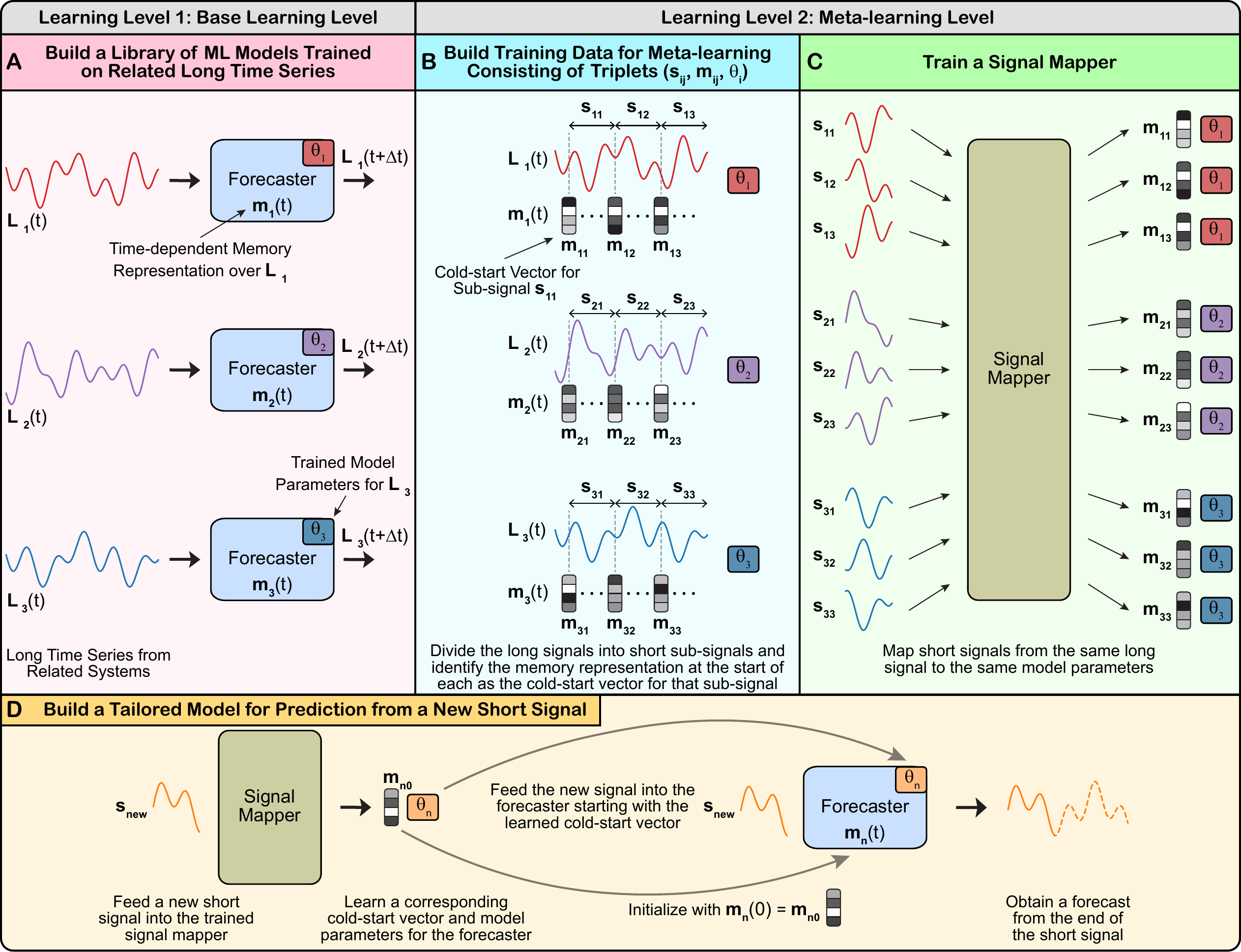}
	\caption{\justifying\textbf{A Schematic of the METAFORS Meta-learning Method.} \textbf{(A)} We train the forecaster separately on each of the available long training series, $\boldsymbol{L}_i$, to construct both a model representation of each corresponding dynamical system, and a cold-start vector at every time-step. \textbf{(B)} We then divide the long signals into short sub-signals, $\boldsymbol{s}_{ij}$, that are the same length as the short new system signals and \textbf{(C)} train the signal mapper to map these short signals to cold-start vectors, $\boldsymbol{m}_{ij}$, appropriate for their start times, and suitable model parameters, $\boldsymbol{\theta}_i$, for the forecaster. \textbf{(D)} Given a short new system signal, $\boldsymbol{s}_{new}$, the signal mapper learns a suitable cold-start vector, $\boldsymbol{m}_{n0}$, and model parameters, $\boldsymbol{\theta}_n$. To make a prediction, we initialize the forecaster with the learned cold-start vector and drive the forecaster with the short new system signal to mitigate errors in the learned cold-start vector. Finally, we evolve the forecaster autonomously with the learned parameters.}
	\label{fig:LARC_Schematic}
\end{figure*}

To address the challenges of data limitations, brittleness, and cold-starting, we introduce Meta-learning for Tailored Forecasting using Related Time Series (METAFORS). METAFORS uses two-level learning (common in meta-learning schemes\cite{Brazdil2022_MetaL_Book}) to build tailored models for data-limited systems of interest by leveraging long time series from other systems suspected to be related.

In our first learning level, we train separate forecasting models (Fig.~\ref{fig:LARC_Schematic}A) for each long training series $\boldsymbol{L}_i(t)$. These models share the same base network architecture, and differ only in their trained model parameters, which we denote by a vector $\boldsymbol{\theta}_i$. We then extract multiple short signals $\boldsymbol{s}_{ij}$ from each long signal and, during the training, we record the state of the model's memory -- represented by the cold-start vector $\boldsymbol{m}_{ij}$ -- immediately prior to inputting $\boldsymbol{s}_{ij}$ (Fig.~\ref{fig:LARC_Schematic}B). These triplets (short signals, trained forecaster parameters, and cold-start vectors) form our training dataset for meta-learning.

In the second level (the meta-learning level) a `signal mapper' ML model learns to map short observation signals in the training dataset to appropriate forecaster parameters and cold-start vectors (Fig.~\ref{fig:LARC_Schematic}C). By leveraging knowledge across related systems, METAFORS generalizes to short signals from new systems without requiring explicit knowledge of governing equations or contextual labels (Fig.~\ref{fig:LARC_Schematic}D).

There are several aspects of METAFORS that differentiate it from related work. 
Previous studies have shown that memory-based ML forecasters, when provided with explicit knowledge of relevant dynamical parameters or contextual labels, can predict the long-term statistics (e.g., climate) of new systems with unseen dynamics, making them valuable for forecasting system behavior under specific parameter regimes, e.g., forecasting climate evolution in response to rising CO\textsubscript{2} levels\cite{Patel2021_NonstationaryRC,Kong2021_MLSystemCollapse,Koglmayr2024_ParamAwareNGRC,Panahi2024_AdaptableRC}. METAFORS addresses a different challenge: rapid generalization to new systems with only a short time series and no explicit contextual indicators or knowledge of dynamical parameters. The meta-learning framework facilitates this context-free generalization. While meta-learning has been explored in forecasting and beyond \cite{Finn17_MAML,Yao2019_HSML,Raghu2019_ANIL,Rusu2019_LEO,Wu2018_MeLA,Joshaghani2023_RetailMetaL,Talagala2023_MetaL_for_Forecasting}, we are aware of only a few studies that focus on few-shot forecasting of dynamical systems \cite{Canaday2021MARC,Kirchmeyer2022a_CoDA,Yin2021_LEADS,Wang2022_DyAd,Panahi2025_CriticalTransitions,Oreshkin2021_MetaL_Zeroshort_TSF}. Among these approaches, METAFORS stands out because it does not rely on specific neural network architectures, both initializes and generalizes the forecaster model (accommodating memory-based models), requires no contextual information or domain knowledge, and needs no re-training when presented with new systems.

The versatility that METAFORS confers to ML models with memory is appealing in part because of existing and potential connections between information processing in artificial and biological neural networks. METAFORS has overlap with Hopfield’s notion of associative memory\cite{Hopfield1982,Kong2024_IndexBasedRC} in that it enables the recall of an entire pattern (full memory initialization) from just a partial input (short signal). It goes beyond associative memory by generalizing previously unseen partial inputs to appropriate memory initializations (cold-start vectors) and ML model parameters for that case. Other related recent work\cite{Lu2020_IGS,Kim2020_RNNChaoticMemories} has employed artificial neural network models with memory for time series forecasting to suggest biologically feasible learning paradigms that capture the flexibility of biological neural systems to learn different tasks simultaneously. These studies advance multi-task learning for time series forecasting by addressing task identification through either pre-processing that separates data distributions \cite{Lu2020_IGS} or by employing a contextual input \cite{Kim2020_RNNChaoticMemories}. METAFORS aims to achieve task flexibility without explicit context awareness or prior separation of tasks, enabling generalization even when data distributions overlap considerably.

The general technique of METAFORS -- combining a short `system signal' from the system of interest with more abundant data from related systems through a two-level meta-learning process -- can be applied broadly to ML approaches with memory. These approaches will differ in the structural representations of both the ML model parameters for the library members and the encoded memory needed for cold starting. However, we do not anticipate these varied representations to present a fundamental challenge for implementing a METAFORS scheme. The method is also applicable to memoryless ML types, for which its implementation simplifies considerably because the cold-start vectors are not needed and the signal mapper has to learn only model parameters.

To demonstrate METAFORS' utility, we employ reservoir computers (RCs)
as our memory-based ML forecasting models. For simplicity, we also use an RC for the signal mapper. RCs have been shown to perform well for data-driven prediction and analysis of dynamical systems\cite{ESANN_Res_Overview,Sun2024_ESN_Review,Lukosevicius2009_RC_Review}, even those whose behavior is complex (e.g., dynamics on extended networks\cite{Srinivasan2022_ParallelRC_for_Networks}) or those that exhibit sensitivity to initial conditions\cite{Lukosevicius2009_RC_Review,Tanaka_RC_Review_2019,Lu_and_Pathak,Pathak_Hybrid,Krishnagopal2020_Sep_of_Chaotic_Signals,Pathak2018_SpatioTemporalChaosRC,Patel2021_NonstationaryRC,Wikner2020_CHyPP,Bollt_RC_VAR} (i.e., “chaotic” systems).  We emphasize, however, that in typical implementations, RCs can also struggle with data-intensity\cite{Lu2018_Attractor_Reconstruction}, brittleness\cite{Goring2024_OutOfDomain,Yan2024_RC_Opps_and_Challenges}, and cold starting\cite{Gauthier2021_NGRC,Zhang2023_Catch22,Lu2018_Attractor_Reconstruction,Grigoryeva2024_ColdStartRC}.

In this work, we show that METAFORS enables ML forecasting to overcome each of these challenges. Using simulated data from well-studied chaotic systems, we demonstrate that METAFORS can build tailored forecasters for systems with unseen and unknown dynamics using only short signals from these systems and no other contextual tags or domain-specific knowledge. We highlight that METAFORS captures both short-term evolution and long-term climate in several important scenarios: when the signals from the new system and the training systems exhibit substantially and qualitatively different dynamics; when the training signals originate from dynamical systems of distinct functional forms; and when the state of the underlying dynamical systems is only partially measured. Moreover, when the available system signals are very short, METAFORS' cold-starting is essential to accurate forecasting, even when the new signal and the training signal(s) are known to have the same underlying dynamics.
\section{Results} \label{sec:Results}
\noindent
We use toy chaotic systems as a controlled experimental setting to test METAFORS' effectiveness. These systems -- the logistic map, the Gauss iterated map, and the Lorenz-63 equations -- are widely studied in nonlinear dynamics because they exhibit rich, complex behaviors despite being governed by relatively simple equations. Their well-characterized dynamics allow us to systematically generate ground truth data, and examine how system-relatedness, signal length, and library data coverage influence METAFORS' ability to construct tailored forecasts. Moreover, these systems transition between qualitatively different behaviors with small changes in their dynamical parameters. This dynamical sensitivity demonstrates an important way in which a system of interest's underlying dynamics may exacerbate the impact of ML model brittleness on forecast quality, and provides challenging conditions in which to test METAFORS' generalization. In our experiments, we refer to the short system signals as `test signals,' and we evaluate METAFORS against ground-truth data from these systems.

For the cases we consider, the data in both the long library signals and short test signals are discrete in time. This discreteness arises either because of sampling in the case of continuous-time systems, where we denote the sampling interval as ${\Delta t}$ (and take the ML prediction time step to also be ${\Delta t}$), or because the system itself evolves in discrete-time, as with the logistic map. For continuous-time systems, ${\Delta t}$ can be short compared to the system's characteristic time scales (e.g., its Lyapunov time). By contrast, for discrete-time systems, the time step is typically on the scale of the system's evolution dynamics. 

The dynamical systems we study here all have attracting sets, or `attractors,' in their state space, towards which trajectories starting within corresponding `basins of attraction' evolve over time. A key challenge in forecasting from short time series is that insufficient sampling of the attractor's dynamics limits the ability to approximate governing functions. More formally, learning a system’s dynamics corresponds to inferring a function -- e.g., ${d\boldsymbol{x}/dt=\boldsymbol{F}_{\boldsymbol{p}}(\boldsymbol{x})}$ for ordinary differential equations or ${\boldsymbol{x}_{n+1}=\boldsymbol{M}_{\boldsymbol{p}}(\boldsymbol{x}_n)}$ for discrete-time systems. METAFORS performs context-free learning of these functions using only observed state sequences ${\boldsymbol{x}(t)}$, without knowledge of system parameters ${\boldsymbol{p}}$, by leveraging information from systems with similar dynamics (i.e, similar functions ${\boldsymbol{F}}$ or ${\boldsymbol{M}}$) for which more data are available.

We first test METAFORS in an idealized setting where both the library and test signals come from the same type of system, the logistic map, and differ only in their parameters and initial conditions. Even in this simple case, where the dynamics follow a quadratic equation, standard machine learning methods that lack explicit knowledge of the governing equations or parameters struggle to generalize effectively. We then extend our analysis to a more challenging scenario, where the library and test signals are drawn from two distinct types of dynamical systems -- the logistic map and the Gauss iterated map. Here, METAFORS must use only the test signal itself to construct a tailored forecasting model, without knowing which system generated the test signal nor any other contextual labels. Finally, using the Lorenz-63 equations, we show that METAFORS can successfully integrate information from both test and library signals to improve forecast accuracy, even when each signal contains only partial information of the state of the system that generated it.

We emphasize that every METAFORS library member consists of a long training signal and its corresponding ML model representation (trained parameters $\boldsymbol{\theta}$) only. To present results in pictorial form and to discuss context/parameter-aware baseline methods, we frequently refer to the \emph{dynamical parameters} used to generate the long library signals and short test signals, e.g., the parameters of the logistic map or Lorenz equations. These dynamical parameters are not, however, included in METAFORS' library and METAFORS can neither access them nor use them to make forecasts.

\subsection{Implementation overview}
\label{subsec:Implementation_Overview}
\noindent
Here, we provide an overview of our reservoir computing implementation of METAFORS and leave a more detailed description to the Methods section.

\begin{figure}[t]
	\centering
	\includegraphics[width=\linewidth]{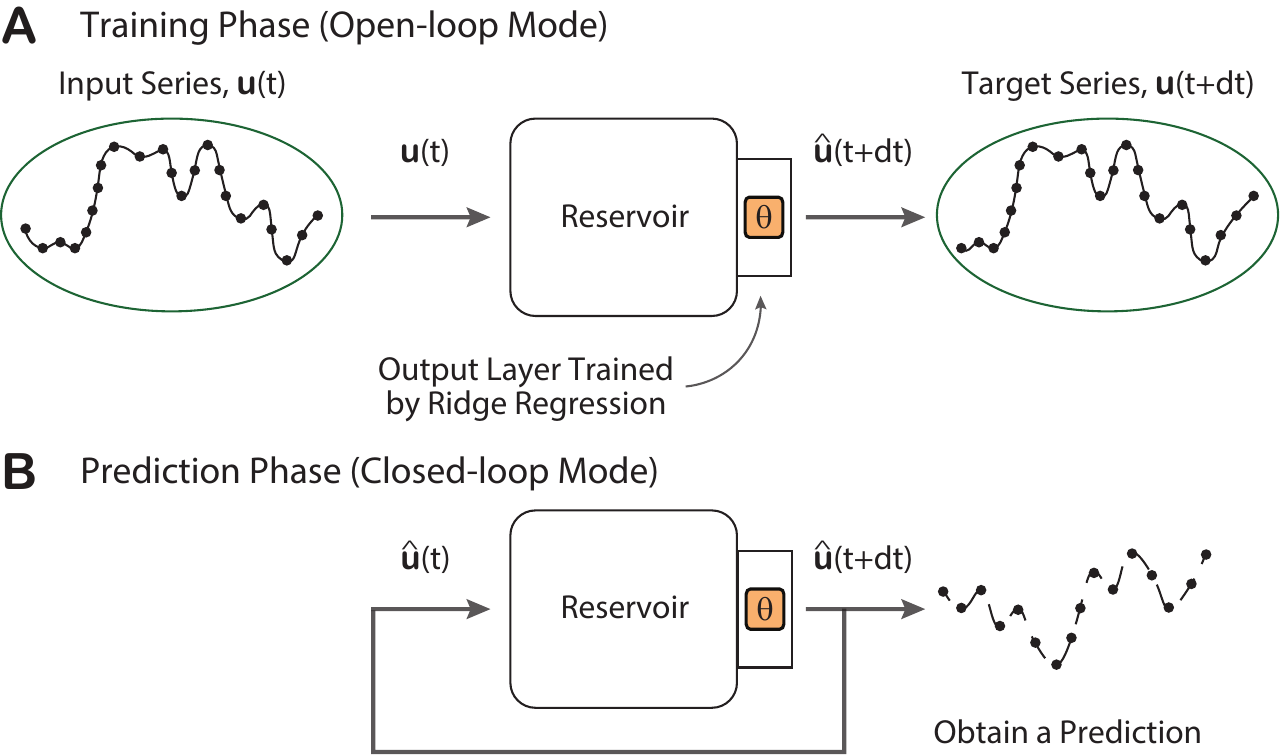}
	\caption{\justifying\textbf{A Reservoir Computer for Time Series Prediction}}
	\label{fig:Res_Diagram}
\end{figure}

The central component of a reservoir computer (RC) is a recurrent neural network with fixed, randomly chosen links (see \cref{subsec:RC_Pred}), referred to as the `reservoir'. Each node/neuron in the reservoir network has an associated continuous-valued activation level, and the collection of all node activations at a given time constitutes the state of the reservoir at that time, ${\boldsymbol{r}(t)}$. The reservoir state evolves over time in response to an external input signal, ${\boldsymbol{u}(t)}$, and is influenced, at every time step, by its own state at the previous time step via directed and weighted interactions between its nodes. This recurrence gives the reservoir `memory.' We generate `outputs,' ${\hat{\boldsymbol{u}}(t+\Delta t)}$, of the reservoir by forming linear combinations of the reservoir nodes' activations, or, equivalently, applying a linear operator -- called the `output layer' -- to the reservoir state vector. In this work, we train a forecaster RC for time series prediction by driving it with a training signal in `open-loop' mode (Fig.~\ref{fig:Res_Diagram}A) and then choosing the output layer, such that the outputs at each time step closely match the target at the next time step, ${\hat{\boldsymbol{u}}(t+\Delta t)\approx\boldsymbol{u}(t+\Delta t)}$. We make this choice by simple linear regression, and use a regularization parameter to prevent over-fitting. To make a prediction, we feed the RC's output back as its input at each time step and it evolves as an autonomous dynamical system in `closed-loop' mode (Fig.~\ref{fig:Res_Diagram}B). The RC's output layer determines the dynamics of this autonomous system and its internal state, ${\boldsymbol{r}(t)}$, determines its position in the state space of those dynamics. Typically, to initialize a forecast, we must synchronize the reservoir state to the test system by driving the RC in open-loop mode with a time series immediately preceding the start of the forecast. If the available time series is too short, the forecast will be inaccurate even if the RC's output layer accurately represents the relevant dynamics. This is because the reservoir state contains memory about previous inputs, so it requires a sufficient `synchronization time' to properly initialize its memory.

We build METAFORS' library of long time series and associated trained ML models as follows. First, using the same reservoir (e.g., with the same random realization of reservoir links), we train a forecaster on each available long signal. Through this process, we obtain, for each $\boldsymbol{L}_i$, using ${\boldsymbol{u}(t)=\boldsymbol{L}_i(t)}$, a well-trained output layer with trained model parameters $\boldsymbol{\theta}_i$ and well-synchronized reservoir states at every time step, which also serve as the cold-start vectors for short signals starting at time $t$, ${\boldsymbol{m}_i(t)=\boldsymbol{r}_i(t)}$.
Next, we extract short signals from the long library signals and construct the training data for the signal mapper, comprising triplets of short signals $\boldsymbol{s}_{ij}$; corresponding RC model parameters $\boldsymbol{\theta}_i$; and cold-start vectors $\boldsymbol{m}_{ij}$ (state of the synchronized reservoir at the start of $\boldsymbol{s}_{ij}$). We then train the signal mapper, which is also an RC in our implementation (but could easily be constructed by other ML schemes), to associate the short signals with their corresponding RC model parameters and cold-start vectors. Once trained, the signal mapper is applied to a short new system signal, ${\boldsymbol{s}_{new}}$ to generate a tailored set of model parameters and a cold-start vector. To mitigate errors in the learned cold-start vector, we then run the forecaster in open-loop mode using ${\boldsymbol{u}(t)=\boldsymbol{s}_{new}(t)}$ with the cold-start vector as the forecaster's initial reservoir state. Finally, the forecaster evolves autonomously from the end of the test signal to generate a forecast.

Once we have obtained a forecast, we can quantify its short-term accuracy by measuring its valid prediction time, which we measure as the earliest time at which the error between the predicted and true trajectories exceeds the standard deviation of the true time series:
\begin{equation}
	T_{valid}=min\bigg{\{}t-N_{test}\Delta t:\bigg{\|}\frac{\hat{\boldsymbol{u}}(t)-\boldsymbol{u}(t)}{ \text{std}(\boldsymbol{u})}\bigg{\|}>1\bigg{\}}.
	\label{eq:Valid_Time}
\end{equation}
Here, the test signal, of $N_{test}$ sequential observations, starts at ${t=0}$ and the prediction starts at ${t_{test}=(N_{test}-1)\Delta t}$. Also, $\hat{\boldsymbol{u}}$ and $\boldsymbol{u}$ are the predicted and true signals, respectively, and $\textrm{std}(\boldsymbol{u})$ denotes the standard deviation of $\boldsymbol{u}$ over time. Both $\textrm{std}()$ and division are performed in a component-wise manner.

We quantify how faithfully a forecast captures the long-term statistics, or climate, of a test system with the autonomous one-step error\cite{Wikner_2022_LMNT}:
\begin{equation}
\epsilon=\left\langle||\boldsymbol{P}\left[\hat{\boldsymbol{u}}\left(t\right)\right]-\boldsymbol{G}\left[\hat{\boldsymbol{u}}\left(t\right)\right]||\right\rangle.
\label{eq:Map_Error}
\end{equation}
Here, ${\hat{\boldsymbol{u}}(t)}$ is the predicted system state at time $t$ and $\langle \boldsymbol{x}\rangle$ denotes the time-average of $\boldsymbol{x}$ over the forecast period. The functions ${\boldsymbol{P}[\boldsymbol{v}]}$ and ${\boldsymbol{G}[\boldsymbol{v}]}$ evolve the system state $\boldsymbol{v}$ at time $t$ forward to time ${t+\Delta t}$ according to the dynamics learned by the forecasting model and those of the true system, respectively; in particular, if $\boldsymbol{u}(t)$ is the true system state at time $t$, ${\boldsymbol{G}[\boldsymbol{u}(t)]=\boldsymbol{u}(t+\Delta t)}$, by definition. The autonomous one-step error quantifies how different the long-term dynamics of the closed-loop RC are from the actual dynamics of the system that generated the data, as measured along the self-generated trajectory that the RC follows in autonomous mode. Since the future states of the reservoir depend only on its own prior outputs rather than external inputs (after a transient following the test signal is discarded), this trajectory reflects the long-term behavior/climate learned by the model. The autonomous one-step error is defined only when the full system state is measured. Thus, while we are most interested in applications where only the partial system state is observed, we also explore cases where full-state measurements are available to facilitate a systematic evaluation of climate replication.

\subsection{Generalizing models of the logistic map with METAFORS}
\label{subsec:Logistic_Results}

\begin{figure*}
	\centering
	\includegraphics[width=\linewidth,scale=1]{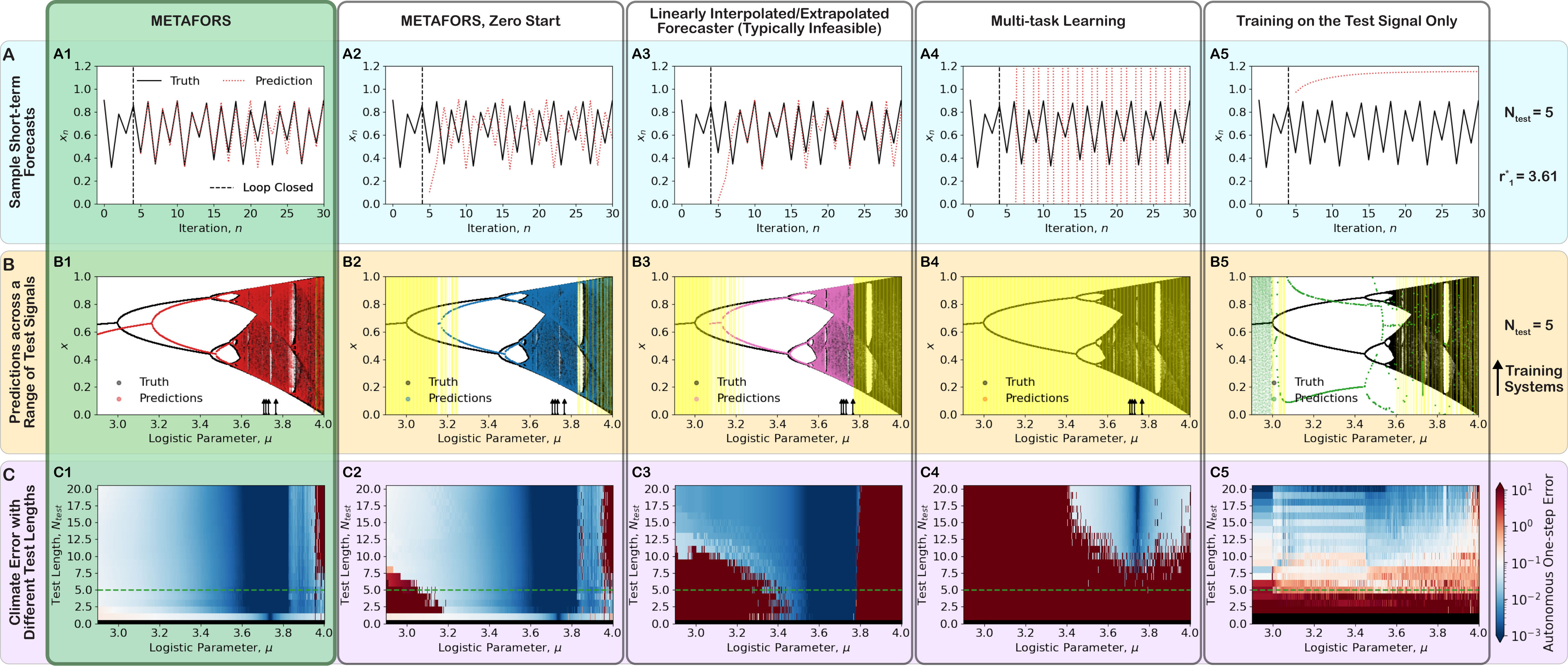}
	\caption{
    \justifying\textbf{Trained on just five stationary signals, METAFORS replicates the logistic map’s dynamics across a large portion of its bifurcation diagram from short test signals with unknown dynamical parameters.} We train METAFORS and baseline approaches on a library of five chaotic logistic map trajectories. The logistic parameters of these library signals, shown as black arrows in \textbf{(B)}, were chosen randomly from ${3.7\leq\mu\leq3.8}$, excluding values with periodic attractor dynamics (\cref{sec:Forecaster_Training_Maps}). \textbf{(A)} Short-term forecasts from a sample short signal containing ${N_{test}=5}$ data points with logistic parameter $\mu_1^*=3.61$. \textbf{(B)}~Bifurcation diagrams constructed with test signals of ${N_{test}=5}$ iterations. Vertical yellow lines indicate values of $\mu$ for which the corresponding forecast leaves the interval $0\leq x\leq1$ and does not return. \textbf{(C)} Median autonomous one-step error over 10 random realizations of the forecaster and signal mapper reservoirs, and of the library signals' initial conditions, for ${1\leq N_{test}\leq 20}$. Green horizontal lines indicate the test length used in panels \textbf{(A)} and \textbf{(B)}. In the true bifurcation diagram \textbf{(B)}, we plot, for each of $500$ evenly-spaced values of ${2.9\leq\mu\leq4}$, the final $500$ iterations of a trajectory of total length $2000$ iterations starting from a random initial condition ${0<x_0<1}$. We start each prediction at iteration $1000$ of the corresponding true trajectory and discard the first $500$ predicted iterations to ensure that any initial transient behaviors do not obscure the forecast long-term climate. We use the subsequent 500 predicted iterations to plot predictions in row \textbf{(B)} and to calculate the autonomous one-step error in row \textbf{(C)}.}
\label{fig:Logistic_Map_Results}
\end{figure*}
\noindent
We employ the logistic map\cite{Lorenz1964_LogisticMap} as a simple testbed 
to demonstrate that METAFORS can generalize our RC forecaster to capture the long-term statistics, or climate, of trajectories with qualitatively different behaviors. The logistic map,
\begin{equation}
	x_{n+1}=\mu x_n(1-x_n)
	\label{eq:Logistic_Map}
\end{equation}
is a simple model of population dynamics. The variable $x$ represents the current population of a species as a fraction of its maximum possible population. Its index, $n$, measures time, typically in years, and ${\Delta n=1}$. When $x$ is small, it grows approximately proportionally to itself at the reproduction rate, or logistic parameter, $\mu$; when $x$ is large, the population declines because resources are scarce. The map is of interest to us, and is commonly studied in dynamical systems, because it exhibits sudden transitions between qualitatively different behaviors as the value of $\mu$ changes, i.e., it is sensitive to changes in its dynamical parameter. Here, we focus on the range ${2.9\leq\mu\leq4}$. For ${2.9\leq\mu<3}$, the map has a single stable fixed point, to which trajectories from all initial conditions in the interval ${0<x_0<1}$ converge. At $\mu=3$, the first bifurcation in a period-doubling cascade occurs as trajectories transition first to a period-2 orbit and then to periodic orbits of increasing period. Chaotic, aperiodic behavior appears beyond ${\mu\approx3.569}$ and is interspersed by windows of periodic orbits as $\mu$ increases through ${3.569\lesssim\mu\leq4}$. We illustrate these bifurcations as the black \textit{Truth} background of the bifurcation diagrams in Fig.~\ref{fig:Logistic_Map_Results}.

We consider each logistic map with a fixed value of $\mu$ as a distinct dynamical system. To demonstrate that METAFORS can capture the climate of unseen test systems with qualitatively different behaviors, we train it on a small number of dynamical systems from the logistic family and show that it can forecast the long-term climate of short test signals with unknown dynamical parameters drawn from a broad section of the map's bifurcation diagram. More precisely, we build a library containing five ML models, each trained on a long trajectory of a chaotic logistic map with logistic parameter drawn randomly from the interval ${3.7\leq\mu\leq3.8}$, excluding values for which the dynamics are periodic. (See \cref{sec:Forecaster_Training_Maps} for details.) We then make predictions with short test signals from $500$ unseen systems with values of $\mu$ evenly-spaced over ${2.9\leq\mu\leq4}$.

We compare the bifurcation diagram constructed by METAFORS to those of four other methods. In \textit{METAFORS, Zero Start}, we use the signal mapper to learn output layer parameters, but no cold-start vector, for the forecaster RC. In this case, we `zero start' the forecaster to make a prediction from a short test signal. That is, we initialize the forecaster's internal state with a zero vector and then drive the forecaster with the test signal to at least partially synchronize its internal state before closing the loop. We also compare to an \textit{Interpolated/Extrapolated Forecaster} (\cref{subsec:Interpolation}) that relies on knowledge of the logistic parameter values for each of the training and test systems. We identify, for each test signal, whether its logistic parameter is within or beyond the range of the library members. If it is within this range, we perform element-wise linear interpolation of the model parameters for the two library members whose logistic parameters most closely bracket those of the test system. Otherwise, we linearly extrapolate from the model parameters of the nearest two library members. Interpolation and extrapolation are not feasible when knowledge of the dynamical parameters of the training and test systems is not available. In our simple \textit{Multi-task Learning} approach (\cref{subsec:Multitask}), we train a single forecaster RC on the union of all long signals in the library. In \textit{Training on the Test Signal Only} (\cref{subsec:Direct_Training}), we train a separate forecaster RC directly on each short test signal. For all of these methods, we zero start the forecaster as in \textit{METAFORS, Zero Start}, to make a prediction.

\begin{figure*}
    \centering
	\includegraphics[width=\linewidth,scale=1]{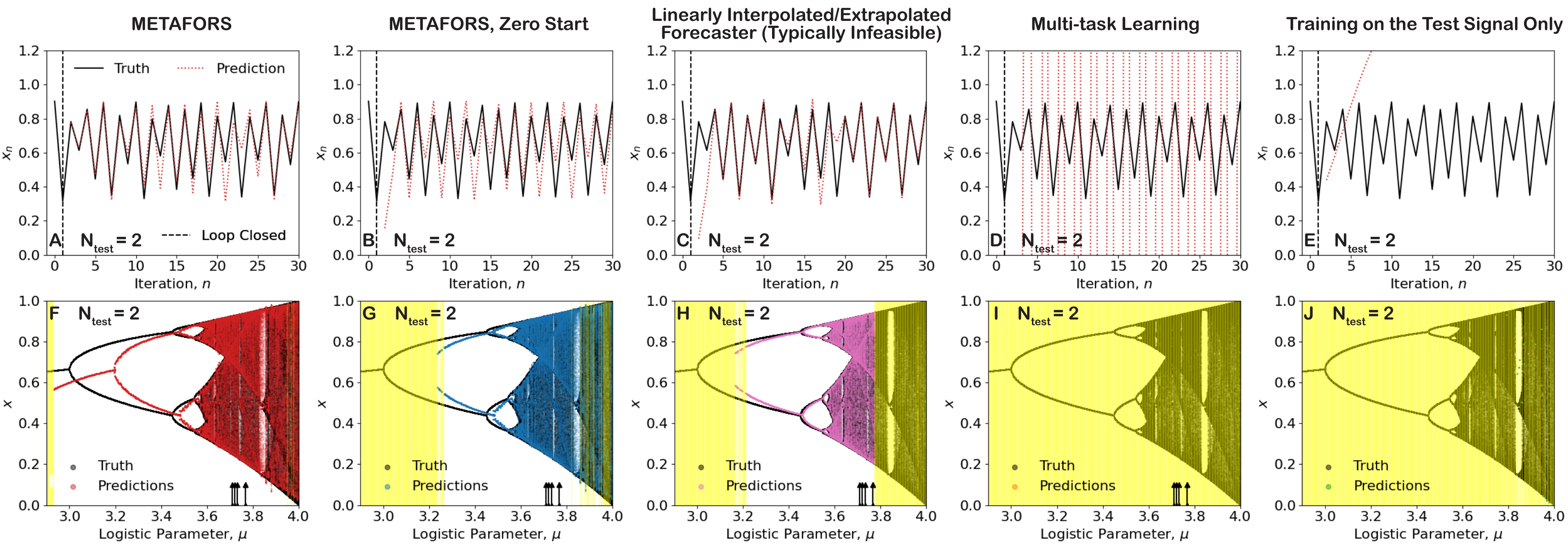}
	\caption{\justifying \textbf{METAFORS replicates the logistic map’s dynamics across a large portion of its bifurcation diagram from test signals with unknown dynamical parameters and containing just ${\mathbf{N_{test}=2}}$ data points.} We train METAFORS on a library of five trajectories from the logistic map with logistic parameters chosen randomly from ${3.7\leq\mu\leq3.8}$ (black arrows, \textbf{F to J}). All signals in the library are chaotic; periodic trajectories are excluded from selection. All test signals contain ${N_{test}=2}$ iterations. \textbf{(A to E)} Example short-term forecasts obtained by METAFORS and baseline methods from a test signal with logistic parameters ${\mu_1^*=3.61}$ and \textbf{(F to J)} bifurcation diagrams constructed by the same methods. Vertical yellow lines indicate values of $\mu$ for which the corresponding forecast leaves the interval $0\leq x\leq1$ and does not return. In the true bifurcation diagram \textbf{(F to J)}, we plot, for each of $500$ evenly-spaced values of ${2.9\leq\mu\leq4}$, the final $500$ iterations of a trajectory of total length $2000$ iterations starting from a randomly chosen initial condition ${0<x_0<1}$. We start each prediction at iteration $1000$ of the corresponding true trajectory and discard the first $500$ predicted iterations to ensure that the forecast long-term climate is not obscured on the plot by any initial transient behavior. We plot the subsequent $500$ predicted iterations.}
	\label{fig:Logistic_Map_Ntest2}
\end{figure*}

In Fig.~\ref{fig:Logistic_Map_Results}(A) we plot short-term forecasts obtained with each of these methods from a short test signal of length ${N_{test}=5}$ with sample value of the logistic parameter ${\mu_1^*=3.61}$. Only the forecast obtained using METAFORS (with cold starting), in Panel (A1), is accurate in the short-term. In Fig.~\ref{fig:Logistic_Map_Results}(B), we use the same methods to construct bifurcation diagrams of the logistic map from test signals of just five iterations (${N_{test}=5}$). Specifically, we generate the test signals by iterating the logistic map  with different values of its bifurcation parameter sampled uniformly over a wide range. We then provide each test signal to the forecasting method of choice and construct the corresponding bifurcation diagram by plotting the long-term predictions of the system's state as a function of the bifurcation parameter $\mu$. These diagrams reveal several noteworthy features. First, Fig.~\ref{fig:Logistic_Map_Results}(B1) shows that METAFORS successfully captures the logistic map's dynamics over a much broader range of logistic parameter values than contained in the library. (We note that METAFORS exhibits strong performance even when the test signal is as short as two iterations -- see Fig.~\ref{fig:Logistic_Map_Ntest2}. We display results for ${N_{test}=5}$ here so as to illustrate some regions where the other methods succeed.) Second, by comparing panels (B1) and (B2), we can see that the cold-start vectors that METAFORS learns are important when the test signals are very short. With test signals of five iterations, only METAFORS' forecasts remain in the unit interval across the whole test set. Both the linearly interpolated/extrapolated forecasters and those constructed by METAFORS without accompanying cold-start vectors (\textit{METAFORS, Zero Start}) replicate the dynamics of the logistic map well over portions of the bifurcation diagram but their predictions completely leave the range of the true trajectories, ${0<x<1}$, for many values of $\mu$. The difference in performance between METAFORS and these methods highlights that cold starting presents a challenge to memory-based forecasters even when the goal is solely to predict a test system's long-term climate rather than its short-term evolution. If not suitably initialized, even a well-trained forecaster may end up in a basin of attraction that is inconsistent with the short test signal. METAFORS mitigates this issue effectively.

We note that because the logistic map is one-dimensional and the training and test signals thus contain full information of the system state, a forecasting model without memory could also perform well for this problem. Here, we use a forecaster RC with memory because we do not, in practice, assume prior knowledge of whether the test and training signals contain observations of the full system state. The logistic map is of interest to us because its dynamical parameter sensitivity allows us to test METAFORS' generalization of the forecaster across test systems with qualitatively different behaviors. Moreover, because the test signals contain information of its full state, we can use the autonomous one-step error (Eq.~\ref{eq:Map_Error}) to measure climate-replication accuracy.

Fig.~\ref{fig:Logistic_Map_Results}(B5) demonstrates that test signals of five iterations are too short for RC forecasters trained directly on the test signals to learn the dynamics of the logistic map well, even though the logistic map is governed only by a simple quadratic equation. In Fig.~\ref{fig:Logistic_Map_Results}(C), we plot the autonomous one-step error of the forecasts from each of our methods with test signals of different lengths. METAFORS captures the climate of test systems with unseen logistic map dynamics more accurately than all of our baseline methods for test signals of length ${N_{test}\leq20}$ iterations. Panel (C4), on the other hand, demonstrates that even when the test signals are long enough to initialize the forecaster well without METAFORS' cold starting, our simple multi-task learning approach to training the forecaster offers good climate replication only when the logistic parameter of the test system is quite close to those of the library members. Once trained by this method, the forecaster's dynamics are independent of the test signal. It can exhibit different climates for different test signals only if the autonomous dynamical system that it forms when operating in the closed-loop mode (Fig.~\ref{fig:Res_Diagram}B) has multiple attractors\cite{Lu2020_IGS}. In our scenario, the multitask forecaster has only one attractor in the interval ${0<x<1}$ that contains the true data; this attractor represents an averaging of the dynamics across the systems seen in training, and the forecaster successfully forecasts within the interval of the true data only in the blue and white portions of Panel (C4). This challenge -- of capturing diverse dynamics with a single trained model -- has also been explored in recent work on parameter-aware forecasting \cite{Patel2021_NonstationaryRC,Kong2021_MLSystemCollapse,Koglmayr2024_ParamAwareNGRC,Panahi2024_AdaptableRC,Kong2024_IndexBasedRC,Kim2020_RNNChaoticMemories}.

Finally, we emphasize that METAFORS is unaware of the logistic parameter values of the training or test signals. While a linearly interpolated/extrapolated forecaster performs comparably to METAFORS once the test signals are long enough, it requires knowledge of the underlying dynamical parameters and is thus typically infeasible in scenarios of interest.

\subsection{Simultaneous generalization with logistic and Gauss maps}
\label{sec:DualLibrary_Results}

\begin{figure*}
	\centering
    \includegraphics[width=\linewidth,scale=1]{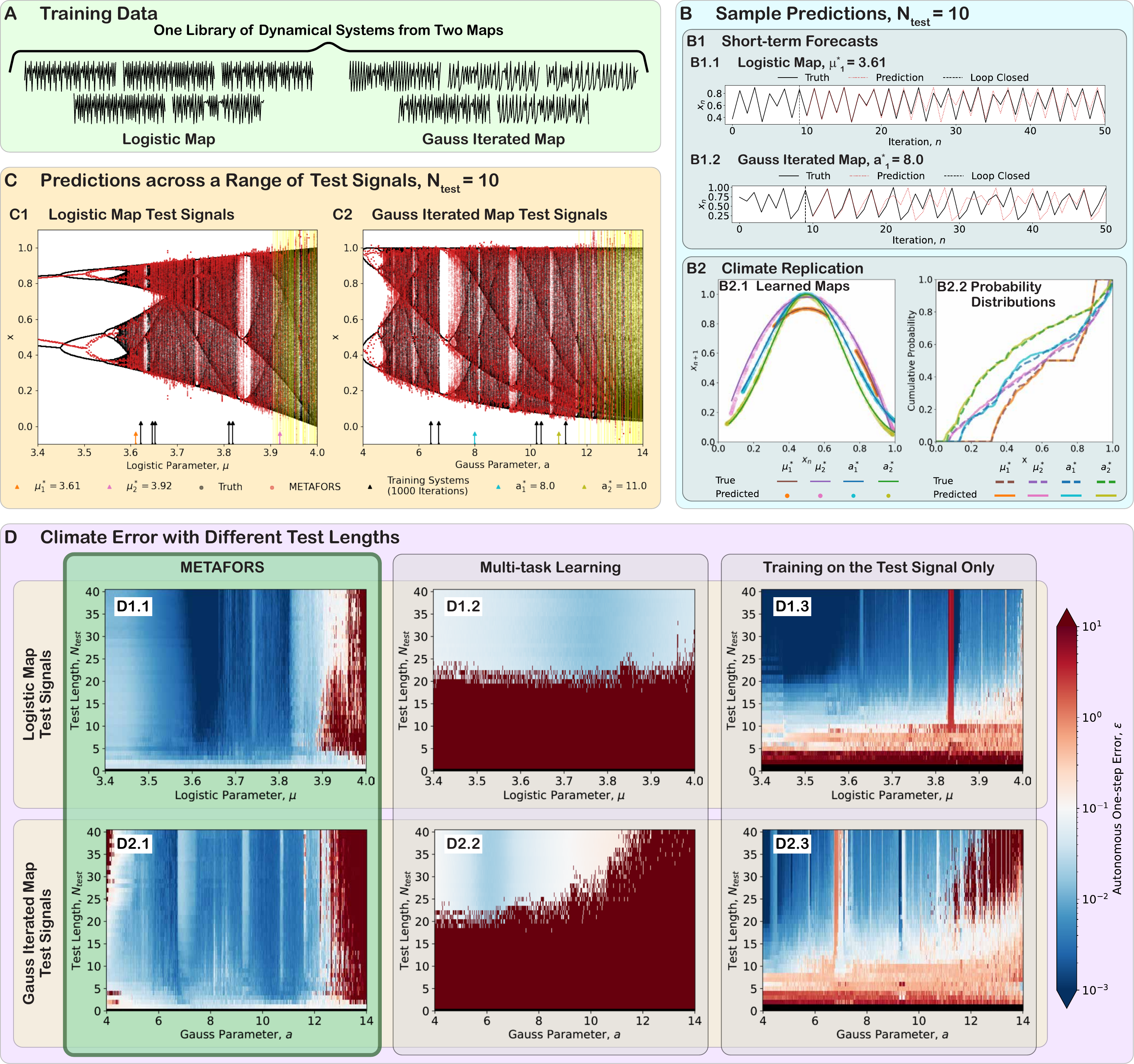}
    \caption{
    \justifying\textbf{METAFORS can learn to represent dynamical systems of distinct functional forms simultaneously.} We train METAFORS on a library of ten chaotic trajectories, each of length 1000 iterations, with dynamical parameters indicated by black arrows in \textbf{(C)}: five logistic map trajectories with logistic parameters chosen randomly from ${3.6\leq\mu\leq3.9}$ and five Gauss iterated map trajectories with exponential parameters chosen randomly from ${6\leq a\leq12}$, excluding values with periodic attractor dynamics. \textbf{(A)} The first 100 iterations of each long library signal. \textbf{(B1)} Short-term forecasts obtained with METAFORS at sample parameter values. \textbf{(B2)} METAFORS' climate replication demonstrated by the one-step update map it learns \textbf{(B2.1)} and the cumulative probability distribution of its predicted trajectories \textbf{(B2.2)} at sample parameter values ${\mu_1^*=3.61}$, ${\mu_2^*=3.92}$, ${a_1^*=8}$, and ${a_2^*=11}$. \textbf{(C)} The true and learned bifurcation diagrams of the logistic map \textbf{(C1)} and the Gauss iterated map \textbf{(C2)}. There are $500$ test signals of length ${N_{test}=10}$ iterations for each map. They are spaced evenly over ${3.4\leq\mu\leq4}$ and ${4\leq a\leq14}$ for the logistic and Gauss iterated maps, respectively. \textbf{(D)} Median autonomous one-step error, calculated over ten random realizations of the forecaster and signal mapper RCs' internal connections and the library signals' initial conditions, for test lengths ${1\leq N_{test}\leq40}$ using METAFORS and baseline approaches. Black regions at ${N_{test}=0}$ for all methods and at ${N_{test}=1}$ for \textit{Training on the Test Signal Only} indicate that no prediction can be obtained. All predicted trajectories are $500$ iterations long.}
\label{fig:Dual_Library}
\end{figure*}
\noindent
METAFORS' performance in Fig.~\ref{fig:Logistic_Map_Results} is aided by the fact that both the library systems and the test systems come from the same one-parameter family of dynamical systems. In this section, we demonstrate that METAFORS can generalize the forecaster to cases where the test signal comes from a system with a different functional form than some of the library systems.

The Gauss iterated map, or mouse map, is given by
\begin{equation}\label{eq:Gauss_Map}
	x_{n+1} = e^{-a(x_n-b)^2}.
\end{equation}
Like the logistic map, its dynamics undergo bifurcations and exhibit qualitatively different behaviors as its parameters, $a$ and $b$, vary.

We demonstrate in Fig.~\ref{fig:Dual_Library} that METAFORS can learn to represent dynamical systems from both the logistic and Gauss iterated maps simultaneously. We train METAFORS on a library of ten long series (Fig.~\ref{fig:Dual_Library}A): five from the logistic map with parameter $\mu$ randomly chosen from ${3.6\leq\mu\leq3.9}$, and five from the Gauss iterated map with $b=-0.5$ and randomly chosen values of the exponential parameter ${6\leq a\leq12}$, excluding periodic trajectories as before. The Gauss iterated map as we use it in our experiments, i.e., Eq.~\ref{eq:Gauss_Map}, differs from its usual form\cite{Hilborn2000_GaussMap} by a translation ${x_n\to x_n-b}$. We use this translated version of the map to make it more challenging to distinguish between trajectories from the logistic and Gauss iterated maps. The translation creates substantial overlap in the distributions of states visited by each map, with trajectories from both confined to the interval ${(0,1)}$ and covering a substantial portion of it. This overlap ensures that METAFORS cannot, for instance, learn to identify with the Gauss iterated map all test signals that have a negative entry.

In Fig.~\ref{fig:Dual_Library}(B), we demonstrate that METAFORS captures both the short-term evolution (B1) and long-term climate (B2) of specific sample systems from the chaotic regimes of each map with test signals of only ${N_{test}=10}$ iterations. Fig.~\ref{fig:Dual_Library}(B2.1) shows that the maps traced out by the true and predicted trajectories from four sample test signals (two from the logistic map and two from the Gauss iterated map) agree closely. Fig.~\ref{fig:Dual_Library}(B2.2) plots the cumulative probability distributions of states visited by the same trajectories. Fig.~\ref{fig:Dual_Library}(B2) illustrates the substantial overlap in the distributions of the two maps. METAFORS nevertheless predicts the state distributions of our sample test systems accurately, and successfully distinguishes between unlabeled test signals from each map with very limited data.

In Fig.~\ref{fig:Dual_Library}(C), we forecast from 500 short test signals, of length ${N_{test}=10}$ iterations, at evenly-spaced values of each map's bifurcation parameter and see that METAFORS can indeed capture the climate of test systems from both maps over a broad range of their bifurcation diagrams. While METAFORS does misplace some features of the bifurcation diagrams -- for example, it predicts that the bifurcation of the logistic map from a period-2 orbit to a period-4 orbit occurs at ${\mu\approx3.5}$ instead of ${\mu\approx3.45}$ -- its reconstructions have broadly similar dynamics. Finally, in Fig.~\ref{fig:Dual_Library}(D) we compare METAFORS' climate replication across both maps' parameter ranges to two other methods -- an RC forecaster trained by multi-task learning and RC forecasters trained directly on the short test signals -- with test signals of different lengths.

We emphasize again that METAFORS has no awareness of the parameters that govern the dynamics of the training or test systems. Moreover, it is not explicitly aware of which map has generated any given signal. METAFORS infers the relevant dynamics from the observed test signal alone. Parameter-aware generalization methods, such as simple interpolation/extrapolation, are not readily applicable to this problem. To use such methods, we would require an additional input channel to indicate which parameter a supplied value measures.

\subsection{Generalization in fully and partially observed Lorenz-63 systems}
\label{subsec:Lorenz_System}
\noindent
While we believe that larger applications like weather forecasting could benefit from the approaches used in METAFORS, here we employ a simplified but widely studied model of atmospheric convection, the Lorenz-63 equations\cite{Lorenz63}, as another testbed for METAFORS. First, we briefly demonstrate that METAFORS' forecasts capture the climate of test systems with Lorenz dynamics unseen in training, and then we use METAFORS to predict the short-term evolution of such systems in a few distinct scenarios. Of particular note, we show that METAFORS can still generalize and cold start the forecaster when only partially-observed states of the training and test systems are available. Partial-state forecasting is required in many applications -- especially those where the system of interest is high-dimensional and only a few variables can be feasibly measured, such as in weather-forecasting and epidemiology.

Lorenz systems are governed by three ordinary differential equations:
\begin{subequations}\label{eq:Lorenz}
	\begin{equation}\label{eq:Lorenz_x1}
		\dot{x}_1=\omega_t[v_1(x_2-x_1)],
	\end{equation}\begin{equation}\label{eq:Lorenz_x2}
		\dot{x}_2=\omega_t[x_1(v_2-x_3)-x_2],
	\end{equation}\begin{equation}\label{eq:Lorenz_x3}
		\dot{x}_3=\omega_t[x_1x_2-v_3x_3].
	\end{equation}
\end{subequations}
Each set of values of the parameters $\omega_t$, $v_1$, $v_2$, and $v_3$ define a unique dynamical system. For our experiments, the long library signals correspond to segments of the attractors of chaotic Lorenz systems with different values of the parameter $v_1$ and the time-scale, $\omega_t$. We hold ${v_2=28}$ and ${v_3=8/3}$ fixed. The factor $\omega_t$ does not appear in the original formulation of the equations, but presents the additional challenge of a varying time scale for the dynamics. Due to this variation, we present results in units of reservoir time steps (${\Delta t = 0.01}$ in units of the differential equations) rather than in units of a characteristic dynamical time scale such as the Lyapunov time, which differs across systems. (The Lyapunov time is the typical amount of time required for the distance between two trajectories that are initially close together to increase by a factor of Euler's number, $e$. It quantifies the time-scale over which a system's chaotic dynamics make prediction impossible\cite{Tel_Gruiz_2006_Dissipative_Chaos}.) For context, however, we note that the Lyapunov time of the Lorenz system with standard parameter values ${\omega_t=1}$, ${v_1=10}$, ${v_2=28}$, and ${v_3=8/3}$ is $\tau_{Lyap}\approx 1.104\approx110\Delta t$.

Full details of the experimental parameters defining METAFORS' library and training scheme for our experiments with Lorenz systems are given in \cref{sec:Forecaster_Training_Lorenz}. An example forecast of a fully-observed Lorenz system using METAFORS is shown in Fig.~\ref{fig:Lorenz_Results}(A). Note that the test signal starts at time ${t=0}$ and the prediction starts at time ${t=t_{test}}$.

\subsubsection{Climate replication and forecasting for unseen Lorenz systems}\label{sec:Lorenz_Generalization}

\begin{figure*}
	\centering
	\includegraphics[width=\linewidth,scale=1]
	{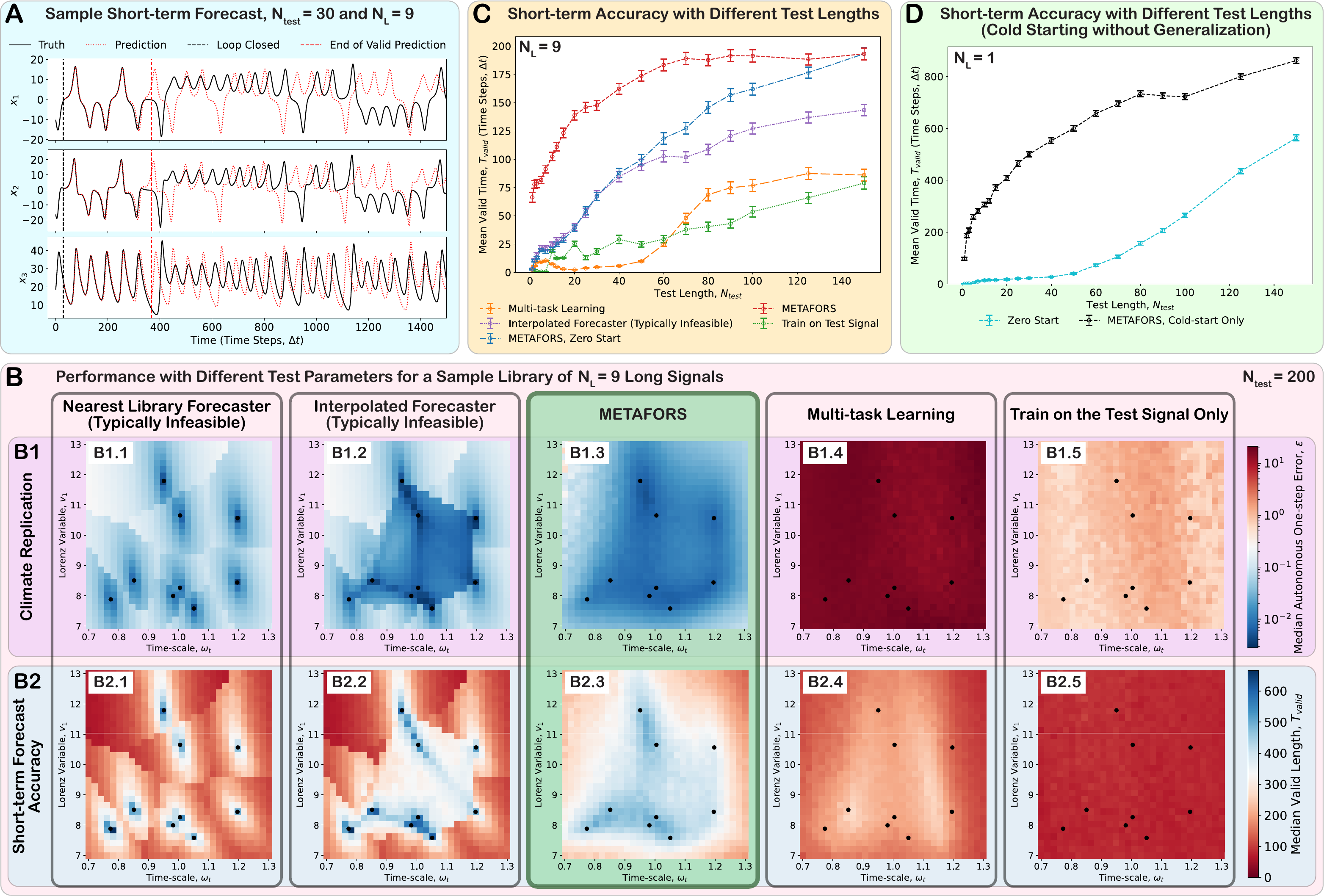}
	\caption{
    \justifying\textbf{METAFORS generalizes the forecaster to unseen fully-observed and partially-observed Lorenz sytems.} \textbf{(A)}~An example METAFORS forecast from a test signal of ${N_{test}=30}$ fully-observed Lorenz states with dynamical parameters ${(\omega_t,v_1)\approx(1.25,10.83)}$. The forecast, starting at ${t_{test}\equiv(N_{test}-1)\Delta t}$ (vertical dashed black line), yields a valid prediction time ${T_{valid}=340\Delta t}$. \textbf{(B)} The dynamical parameters used to generate the nine long library signals (black dots) for panels (A) to (C) were randomly selected from the uniform distributions ${\omega_t\in U[0.75, 1.25]}$ and ${v_1\in U[7.5, 12.5]}$. \textbf{(B1)} Median autonomous one-step error calculated over $500$ full-state forecasts, each of duration $3000\Delta t$. \textbf{(B2)} Median valid times for the same set of forecasts. The test length, ${N_{test}=200}$, is longer than the memory of the forecaster, so that the cold-start vectors learned by METAFORS offer no advantage over the other methods. \textbf{(C and D)} Mean valid prediction time against test signal length with partially-observed training and test systems (only the Lorenz $x_3$-variable is measured) in two distinct scenarios. Error bars denote the standard error of the mean. \textbf{(C)} The dynamical parameters of the $625$ test signals are arranged in a $25\times25$ rectangular grid spanning ${0.7\leq \omega_t\leq1.3}$ and ${7\leq v_1\leq13}$. \textbf{(D)} The library contains only one long time series (${N_L=1}$). This single training time series and all $625$ test signals have different initial conditions but the same dynamical parameters, ${\omega_t=1}$ and ${v_1=10}$. No generalization to new dynamics is required, so we train the forecaster parameters on the sole library signal directly (for both methods) and METAFORS' signal mapper learns only a cold-start vector, ${\boldsymbol{r}(0)}$, for the forecaster.
    }
    \label{fig:Lorenz_Results}
\end{figure*}
\noindent
In Fig.~\ref{fig:Lorenz_Results}(B), we present results for a library constructed from nine fully-observed Lorenz systems with different parameter values randomly chosen from the ranges ${0.7\leq\omega_t\leq1.3}$ and ${7\leq v_1\leq13}$, illustrating how the long-term climate replication (B1) and short-term forecast accuracy (B2), measured by the autonomous one-step error and valid prediction time, respectively, vary with the test parameters. We compare METAFORS' generalization ability to a few simple baseline approaches. First, for the \textit{Nearest Library Forecaster}, we rescale the dynamical parameters $\omega_t$ and $v_1$ used to generate the long library signals such that they span a unit interval along both axes. Then, for each test signal, we make a prediction using the forecaster RC of the long library signal whose re-scaled dynamical parameters are nearest to those of the test system. For the \textit{Interpolated Forecaster} (\cref{subsec:Interpolation}), we perform linear interpolation of the forecaster model parameters if the test system lies within the convex hull of the library. Otherwise, we use the forecaster RC of the nearest library member. We emphasize that in typical applications we do not know the dynamical parameters associated with either the short test signals or the long library signals. So, both this method and \textit{Nearest Library Forecaster} are typically infeasible. We include them for comparison to schemes that rely on additional information (i.e., the typically unknown dynamical parameters). Since no cold-start vector is learned in either of these methods, in our simple multi-task learning approach, or when training a forecaster on the test signal directly, we `zero start' these models. That is, we obtain forecasts by synchronizing the forecaster to the test signal from a zero-vector internal state, $\boldsymbol{r}(0)=\boldsymbol{0}$, before prediction begins at ${t=t_{test}}$.

\begin{figure*}
	\centering
	\includegraphics[width=\linewidth,scale=1]{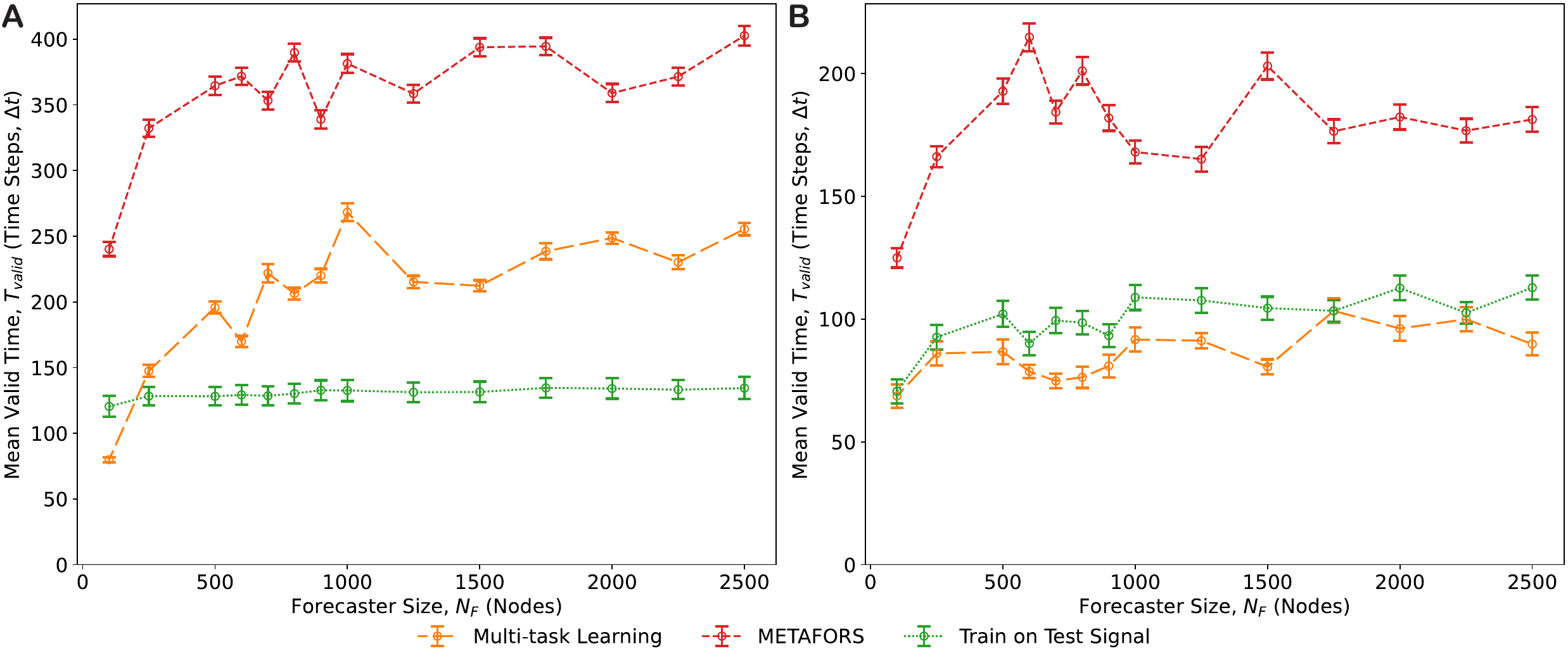}
	\caption{\justifying \textbf{Even with a more powerful forecaster (with more nodes), our simple multi-task learning approach cannot match METAFORS' performance.} We plot mean valid prediction time, $T_{valid}$, against forecaster size, $N_F$, with \textbf{(A)} fully-observed Lorenz systems and \textbf{(B)} partially-observed Lorenz systems ($x_3$-only). In \textbf{(A and B)}, the test signals have ${N_{test}=200}$ sequential data points (long enough that METAFORS' cold starting of the forecaster offers no advantage over the other methods). The library consists of the same ${N_L=9}$ long signals whose dynamical parameters are marked as black dots in Fig.~\ref{fig:Lorenz_Results}(B) and we calculate mean valid times over a test set of $625$ time series arranged in a $25\times25$ rectangular grid spanned by ${7\leq v_1\leq13}$ and ${0.7\leq\omega_t\leq1.3}$, as in \cref{sec:Forecaster_Training_Lorenz}. Error bars denote the standard error of the mean.}
	\label{fig:Valid_Time_vs_ForecasterSize}
\end{figure*}

Fig.~\ref{fig:Lorenz_Results}(B1.1) and (B2.1) illustrate the brittleness of the RC forecasting models in our library. Each forecaster model in the library provides strong climate replication (B1.1) and accurate short-term forecasts (B2.1) only if the test signal parameters are very close to those of the system on which it was trained. Panel (B1.4), on the other hand, demonstrates that our basic multi-task learning approach to training the forecaster fails to capture the climate of unseen test systems even when their dynamical parameters are quite close to those used in the library. The multi-task learning approach is slightly more useful for short-term forecasting (B2.4), but struggles in a way that is common to multi-task learning methods: by training to improve performance on the library members generically, it does not forecast any single system accurately. We show in Fig.~\ref{fig:Valid_Time_vs_ForecasterSize} that a larger (i.e., more powerful) multi-task forecaster RC does not close the performance gap between the basic multi-task learning method and METAFORS. Panels (B1.3) and (B2.3) demonstrate that METAFORS facilitates the generalization that is required for this problem. Over a wide range of test signal dynamics, METAFORS offers lower autonomous one-step errors (B1.3) and higher valid prediction times (B2.3) than the other methods. In particular, METAFORS offers longer valid prediction times than does the parameter-aware \textit{Interpolated Forecaster} (B2.2) method unless the test dynamics are very similar to the dynamics of one of the library members.

\begin{figure*}
	\centering
	\includegraphics[width=\linewidth,scale=1]{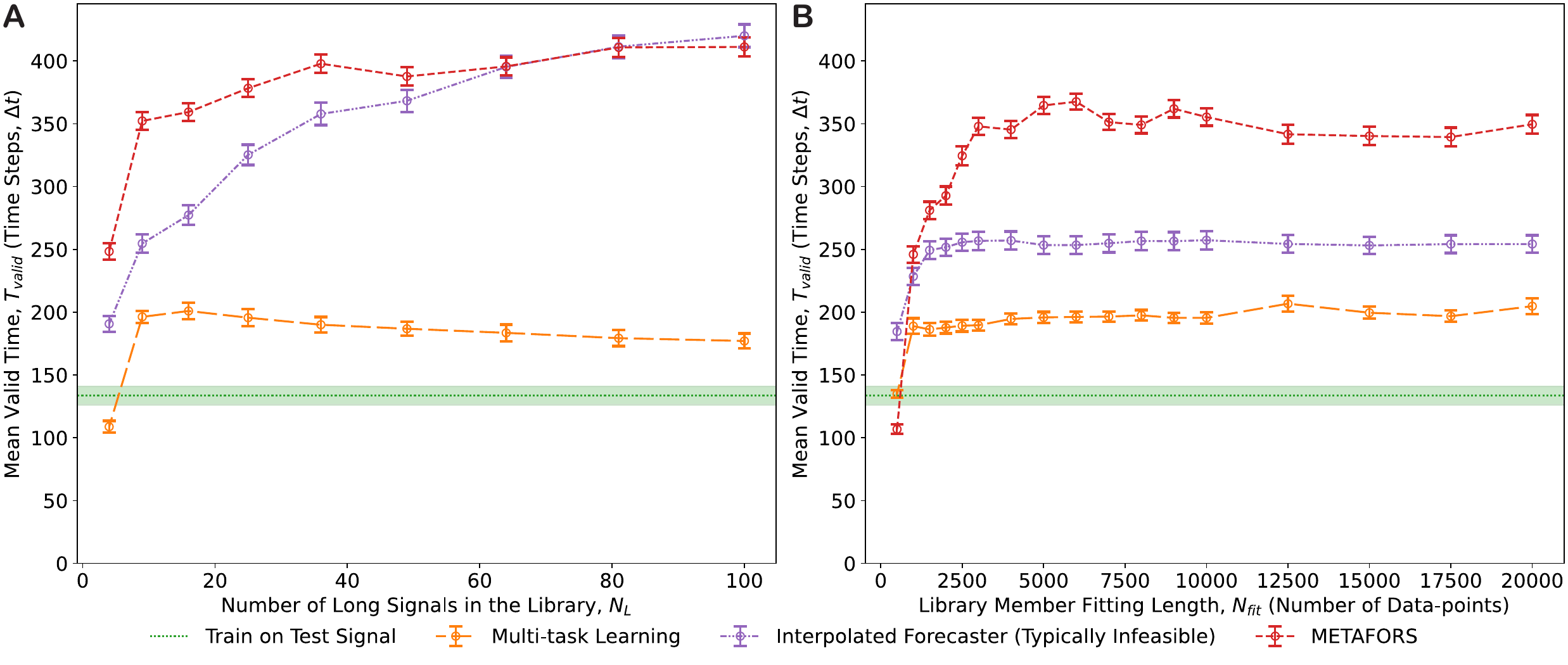}
	\caption{\justifying \textbf{The effect of library structure (length and number of long signals) on generalization performance.} In Fig.~\ref{fig:Lorenz_Results}(B) we show how the climate-replication performance of METAFORS and a few baseline forecasting methods depends on the relationship between the dynamical parameters of the test and library signals for a sample library containing $N_L=9$ long signals of fixed length ${N_{train}=6000}$ (${N_{fit}=5000}$). Here, explore how the structure of the library affects the expected short-term prediction quality over a range of dynamical parameters of the test signals. Namely, we plot mean valid prediction time against \textbf{(A)} the number of library members, $N_L$, and \textbf{(B)} the number of data-points in each long library signal used to fit the forecaster's trainable parameters, $N_{fit}$. In \textbf{(A)}, we fix ${N_{fit}=5000}$. In \textbf{(B)}, we fix ${N_L=9}$. In \textbf{(A and B)}, the test signals are of fixed length  ${N_{test}=200}$ and the library contains $N_L$ long signals with Lorenz parameters chosen randomly from the uniform distributions ${v_1\in U[7.5, 12.5]}$ and ${\omega_t\in U[0.75,1.25]}$. We calculate mean valid times over a test set of $625$ time series arranged in a $25\times25$ rectangular grid spanned by ${7\leq v_1\leq13}$ and ${0.7\leq\omega_t\leq1.3}$, as in \cref{sec:Forecaster_Training_Lorenz}. Error bars and shaded regions indicate the standard error of the mean. Independent of the fitting length, $N_{fit}$, we discard a transient of ${N_{trans}=1000}$ data points at the start of each long library signal to train the forecaster RC. Note that the \textit{Interpolated/Extrapolated Forecaster} requires knowledge of the Lorenz parameters $v_1$ and $\omega_t$ governing the dynamics of the library and test signals.}
	\label{fig:Valid_Time_vs_Lib_Properties}
\end{figure*}

Fig.~\ref{fig:Valid_Time_vs_Lib_Properties} explores further how the library structure (i.e., the length and number of library signals) affects METAFORS' performance.

\subsubsection{With very short test signals, proper cold starting is essential}
\label{sec:Cold_Start_and_Generalization}
\noindent
We plot mean valid prediction time against test signal length for METAFORS and a few baseline methods in Fig.~\ref{fig:Lorenz_Results}(C). Here, both the training and test signals contain only partially-observed Lorenz states (the $x_3$-component only). We present similar results with fully-observed Lorenz systems in Fig.~\ref{fig:TValid_vs_NTest_FullyObserved}(A). METAFORS' ability to cold start forecasts strikingly increases valid prediction times when the test signals are short. With test signals consisting of ${N_{test}=20}$ data points, for instance, METAFORS' mean valid prediction time, ${T_{valid}\approx139\Delta t}$, is just over seven times the length of the test signals. All other methods synchronize the forecaster RC to the test signal from a zero-vector internal state ${\boldsymbol{r}(0)=\boldsymbol{0}}$ (zero starting) and thus cannot offer comparably long valid times until the duration of the test signal is similar to that of the forecaster's memory. When the test signals are long enough that cold starting is not required, METAFORS' valid prediction time still settles at a higher plateau value than even our parameter-aware \textit{Interpolated Forecaster} method, the best performing of the baseline approaches we consider. The comparably poor performance of forecaster RCs trained directly on the test signals (\textit{Train on Test Signal}), in contrast, highlights that even these longest test signals are much shorter than is required to train an accurate RC forecaster.

\subsubsection{Effective cold starting when generalization isn't required}
\label{sec:Cold_Start_Only}
\noindent
While we typically do not expect ML forecasting models trained on data from just one dynamical system to generalize well to test signals with different dynamics, we do expect that a forecaster should offer useful predictions for test signals that exhibit essentially the same dynamics as those of its training system. ML forecasters with memory, however, still struggle in this scenario unless the test signal is sufficiently long to initialize their memory.

\begin{figure*}
	\centering
	\includegraphics[width=\linewidth,scale=1]{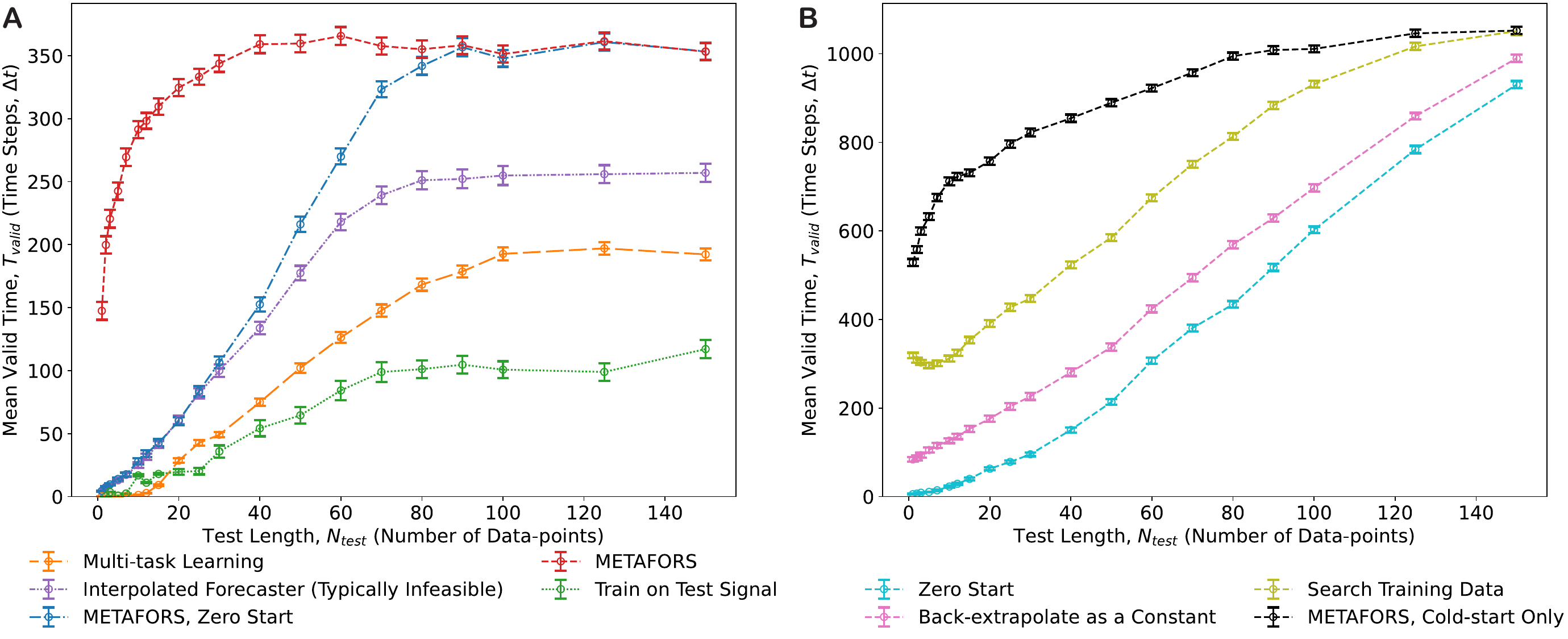}	\caption{\justifying \textbf{Model generalization and cold-starting performance against test signal length in fully-observed Lorenz systems.} \textbf{(A)} Using a library comprising $N_L=9$ long signals with Lorenz parameters indicated by the black dots in Fig.~\ref{fig:Lorenz_Results}(B) we calculate mean valid times over $625$ test signals arranged in a $25\times25$ rectangular grid spanning the space defined by ${0.7\leq \omega_t\leq1.3}$ and ${7\leq v_1\leq13}$. For \textit{METAFORS, Zero Start} (blue), the signal mapper RC learns model parameters but no cold-start vector for the forecaster. For that method and all others except METAFORS, we zero start the forecaster: we synchronize it to the test signal from a zero-vector internal state, ${\boldsymbol{r}(0)=\boldsymbol{0}}$, and then predict in autonomous/closed-loop mode (Fig.~\ref{fig:Res_Diagram}B) from the end of the test signal. \textbf{(B)} We use a library comprising ${N_L=1}$ long time series. This single training signal and all $625$ test signals are segments of the attractor for the standard Lorenz system, ${\omega_t=1}$ and ${v_1=10}$, with different initial conditions. Since there is only one library member, the signal mapper learns only a cold-start vector, ${\boldsymbol{r}(0)}$, for the forecaster. For all methods, we train the forecaster parameters on the library signal directly. In \textbf{(A and B)}, the training and test signals contain fully-observed Lorenz states and error bars denote the standard error of the mean.}
	\label{fig:TValid_vs_NTest_FullyObserved}
\end{figure*}

In Fig.~\ref{fig:Lorenz_Results}(D), we train METAFORS on a single long library signal with standard Lorenz parameters ${\omega_t=1}$ and ${v_1=10}$ and test it on short signals with the same dynamics starting from $625$ different initial conditions. Since there is just one training time series, the signal mapper only has to map short signals to cold-start vectors, with the forecaster's model parameters determined by training on the sole long library signal. In both the partially-observed case (Fig.~\ref{fig:Lorenz_Results}(D)), where we include only the $x_3$-component of the Lorenz system in the training signal and the test signals, and the fully-observed case (Fig.~\ref{fig:TValid_vs_NTest_FullyObserved}(B)), METAFORS' signal mapper enables simple cold-starting of the forecaster. We emphasize that the signal mapper requires no more training data than is traditionally required for forecasting of stationary dynamics; it is trained from the same data as the forecaster in order to cold start predictions at new points in state space.

\begin{figure*}
	\centering
	\includegraphics[width=.5\linewidth,scale=1]{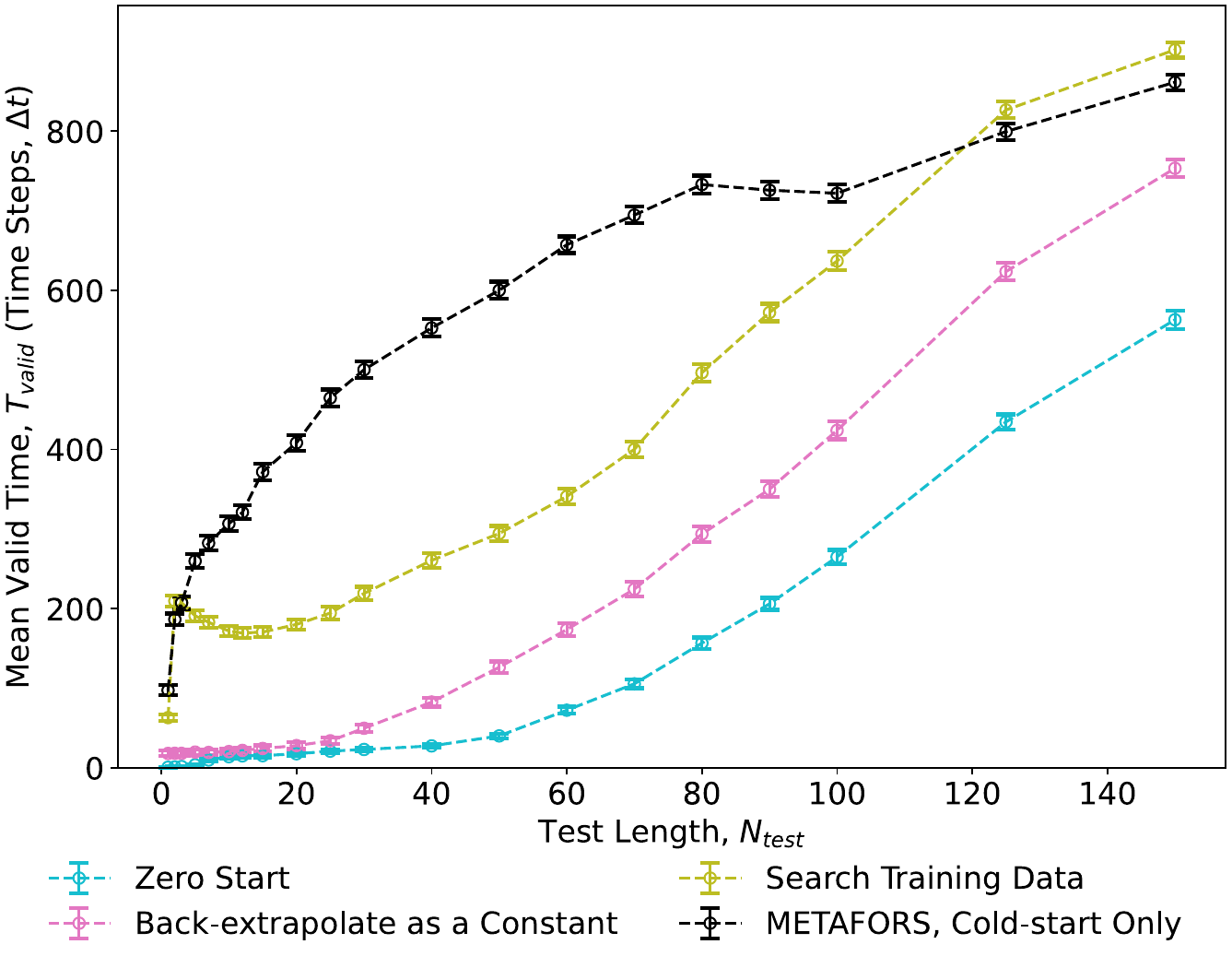}
    \captionsetup{singlelinecheck=off}
    \caption{\justifying \textbf{METAFORS is useful for cold starting even when generalization is not required.} Using a library comprising ${N_L=1}$ long time series, this single training signal and all $625$ test signals are segments of the attractor for the standard Lorenz system, ${\omega_t=1}$ and ${v_1=10}$, with different initial conditions. Since there is only one library member, the signal mapper learns only a cold-start vector, ${\boldsymbol{r}(0)}$, for the forecaster. For all methods, we train the forecaster parameters on the library signal directly. The training and test signals are partially-observed, containing only the $x_3$ Lorenz variable. Error bars denote the standard error of the mean. In \textit{Backward Extrapolation as a Constant}, we extrapolate the test signal backwards from its initial value, ${\boldsymbol{s}_{test}(0)}$, as a constant for two-hundred time steps and then synchronize the forecaster to this extrapolated signal \[\boldsymbol{s}_{test}^{extrap}(t)=\begin{cases}\boldsymbol{s}_{test}(0), & -200\Delta t\leq t<0\\ \boldsymbol{s}_{test}(t), &  0\leq t\leq t_{test}\end{cases},\] starting from ${\boldsymbol{r}(-200\Delta t)=\boldsymbol{0}}$. In \textit{Training Data Search}, we search the training series for the segment, $s_{match}$, that minimizes its root-mean-square distance from the test signal. Denoting by ${\boldsymbol{r}_{train}(t_{match})}$ the constructed cold-start vector at the time step of the training signal at which $s_{match}$ begins, we then synchronize the forecaster to the test signal starting from the initial state ${\boldsymbol{r}(0)=\boldsymbol{r}_{train}(t_{match})}$. Searching the library for the closest match to the test signal offers considerable improvement over the two other elementary approaches that we highlight, but it requires a computationally expensive search each time we wish to make a new prediction. Its performance may also depend on how well the training signal covers its associated attractor. METAFORS, on the other hand, can learn a cold-start vector cheaply for each test signal -- no retraining of the signal mapper is required.}
    \label{fig:TValid_vs_NTest_Partial_ColdStart_Sup}
\end{figure*}

The utility of METAFORS' cold starting in this simplified setting is highlighted by comparison to common elementary approaches to initializing a memory-based ML forecaster. We compare METAFORS' performance to a number of alternative initialization methods in Fig.~\ref{fig:TValid_vs_NTest_Partial_ColdStart_Sup}, but focus here (Fig.~\ref{fig:Lorenz_Results}D) on one typical approach, zero starting.

\begin{figure*}
	\centering
	\includegraphics[width=\linewidth,scale=1]{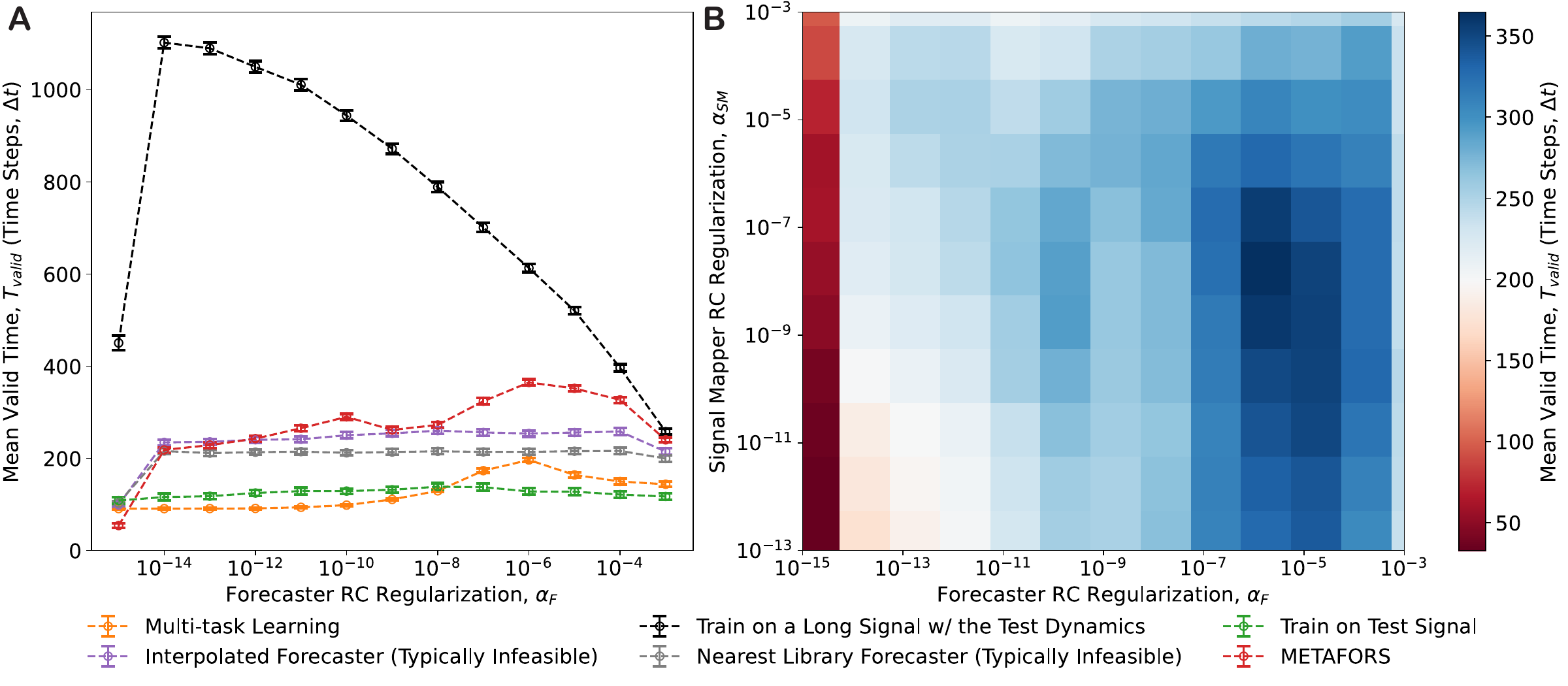}
	\caption{\justifying \textbf{The dependence of generalization performance on Tikhonov regularization strength in fully-observed Lorenz systems.} \textbf{(A)} We plot the mean valid time achieved by METAFORS and each of our baseline methods as we vary the regularization strength used to train the forecaster RC. Error bars denote the standard error of the mean. \textbf{(B)} We plot METAFORS' mean valid prediction time as both the forecaster regularization and the signal mapper regularization vary. In \textbf{(A)} and \textbf{(B)}, we calculate mean valid times over a test set of $625$ time series arranged in a $25\times25$ rectangular grid spanned by ${7\leq v_1\leq13}$ and ${0.7\leq\omega_t\leq1.3}$, as in \cref{sec:Forecaster_Training_Lorenz}. The ${N_L=9}$ long library signals consist of fully-observed Lorenz signals with Lorenz parameters depicted as black dots in Fig.~\ref{fig:Lorenz_Results}(B). The test signals, of $N_{test}=200$ sequential observations, are long enough to synchronize the forecaster's reservoir state well such that its initialization does not matter. The line labeled \textit{METAFORS} in plot \textbf{(A)} is a horizontal slice along the line ${\alpha_{SM}=10^{-8}}$ of the heatmap in plot \textbf{(B)}. With our reservoir computing implementation of METAFORS (and our chosen reservoir hyperparameters, ~Table~\ref{tab:Reservoir_Parameters}), METAFORS' generalization benefits from a higher regularization than forecasting fixed dynamics. When training a forecaster RC optimally for prediction of test systems with identical dynamics to the training system, a lower regularization strength ($\alpha_F^{Trad}\approx10^{-13}$) allows for an excellent fit to the training dynamics, and for high valid prediction times. When generalization to new dynamics is our goal, a higher regularization ($\alpha_F^{META}\approx10^{-6}$) prevents over-fitting to the training systems but limits peak performance. Forecaster regularization is thus an important consideration in implementing a METAFORS scheme. The discrepancy that we identify in our Lorenz experiments may not be universal, but demonstrates that regularization strengths that typically offer good performance when generalization is not required may be suboptimal when generalization is necessary.
    }
	\label{fig:Regularization_Searches}
\end{figure*}

\begin{figure*}
	\centering
	\includegraphics[width=.5\linewidth,scale=1]{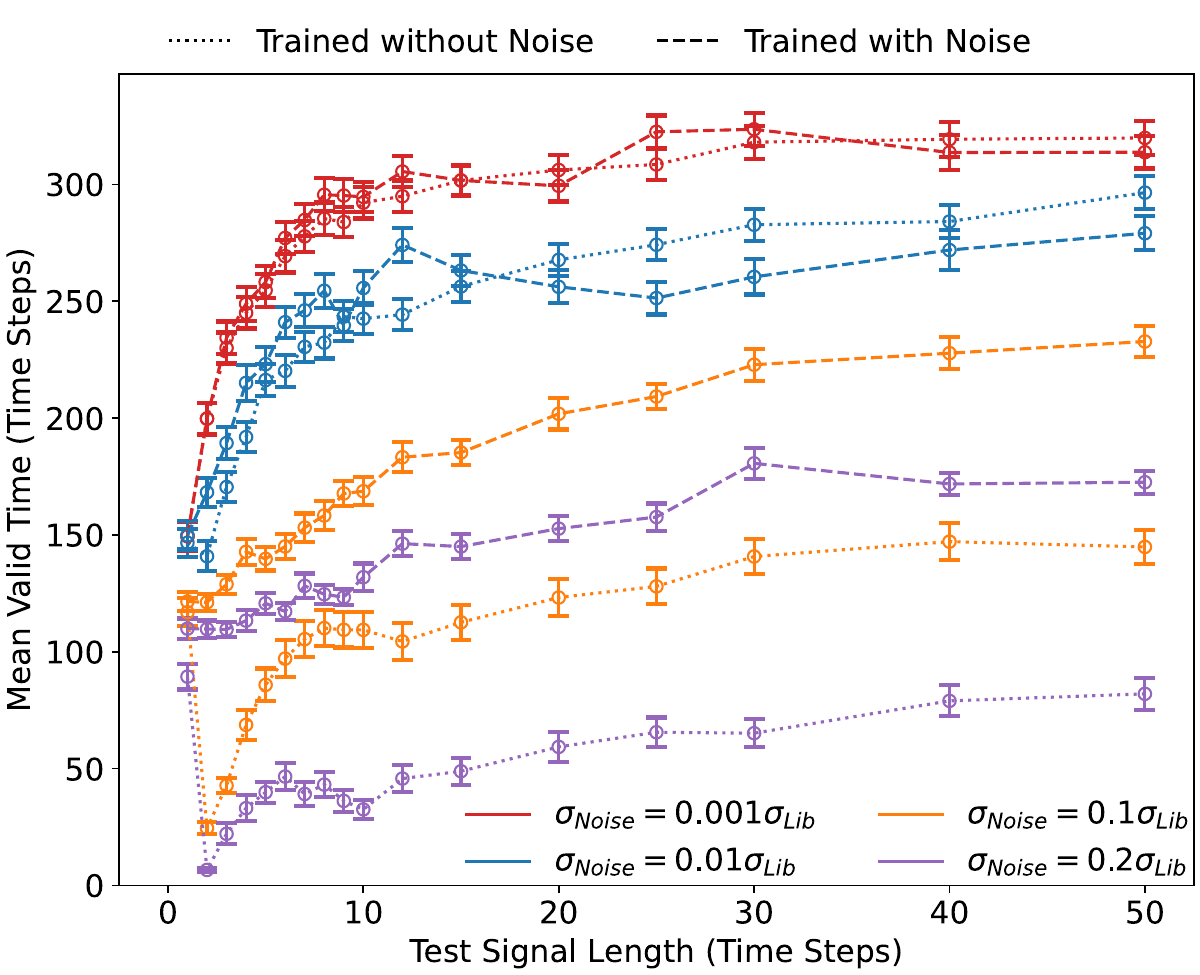}
	\caption{\justifying \textbf{METAFORS is robust to small amounts of noise.} We plot METAFORS' mean valid prediction time as a function of test signal length and noise-amplitude over 625 test signals with dynamical parameters arranged in a $25\times25$ rectangular grid spanned by ${7\leq v_1\leq13}$ and ${0.7\leq\omega_t\leq1.3}$, as in \cref{sec:Forecaster_Training_Lorenz}. Error bars denote the standard error of the mean. We add independent and identically distributed Gaussian observational noise with standard deviation, $\sigma_{Noise}$, in amplitude to each test signal before providing it to METAFORS. $\sigma_{Noise}$ is expressed in multiples of the standard deviation in amplitude of all library members, $\sigma_{Lib}$, with both calculated component-wise. Dotted lines: we train METAFORS on the same noiseless library of nine fully-observed long Lorenz signals whose dynamical parameters are depicted as black dots in Fig.~\ref{fig:Lorenz_Results}(B). Dashed lines: we train METAFORS on the same long library signals, but with observational noise of equal amplitude to that of the test signals. In both cases, we measure valid prediction times against noiseless truth signals. Training on signals with noise of equal amplitude to that of the test signals represents the common scenario that the training and test data are both generated by the same process that is noisy or imperfectly measured. When the amplitude of noise in the test signals is low (${\sigma_{Noise}\lesssim0.01\sigma_{Lib}}$), METAFORS' performance is strong independent of whether it has been trained on data that also contain noise. If the short test signals contain more significant noise (${\sigma_{Noise}\gtrsim0.1\sigma_{Lib}}$), METAFORS performance remains robust so long as it has been trained on data with similar levels of noise.}
	\label{fig:Noise_Plot}
\end{figure*}

We make two brief technical observations. METAFORS' peak valid prediction time in Fig.~\ref{fig:Lorenz_Results}(C) is substantially lower than it is in Fig.~\ref{fig:Lorenz_Results}(D), where we do not require generalization to new dynamics. This discrepancy is related in part to training regularization (Fig.~\ref{fig:Regularization_Searches}). In both the partially-observed (Fig.~\ref{fig:Lorenz_Results}C and D) and fully-observed (Fig.~\ref{fig:TValid_vs_NTest_FullyObserved}) cases, METAFORS offers useful forecasts with test signals of just one data point (the point in state space from which the forecast should start). While cold starting from such limited data is noteworthy, these test signals contain no noise. In Fig.~\ref{fig:Noise_Plot}, we demonstrate that our RC implementation of METAFORS is robust to small amounts of observational noise in the test signals.
\section{Discussion}\label{sec:Discussion}
\noindent
This paper introduces Meta-learning for Tailored Forecasting using Related Time Series (METAFORS), a framework designed to address key challenges in applying traditional ML approaches to forecasting dynamical systems, specifically their dependence on abundant system-specific training data and their brittleness when generalizing to related but distinct systems. By leveraging a library of machine learning models trained on related systems with ample data, METAFORS constructs and initializes tailored forecasting models for unseen systems using only limited observations and no additional contextual information. Our study exhibits METAFORS' capabilities in multiple families of nonlinear systems, demonstrating its robustness, versatility, and potential applicability to real-world systems.

METAFORS offers several key features: {\bf{\emph{(1)~Generalization Across Systems:}}} It forecasts test systems with dynamics that are related to but substantially different from those in the training library, without requiring contextual awareness. {\bf{\emph{(2)~Cold Starting Forecasts:}}} It enables memory-based models to generate accurate forecasts from minimal initialization data, outperforming baselines -- an essential capability for data-limited applications. {\bf{\emph{(3)~Flexibility in Model Architecture:}}} METAFORS is not restricted to specific ML architectures or training schemes; while we employ reservoir computing, the framework is adaptable to other implementations. {\bf{\emph{(4)~Capturing Both Short-term and Long-term Behaviors:}}} It predicts both short-term dynamics and long-term statistical properties (climate), making it applicable to a wide range of forecasting tasks. {\bf{\emph{(5)~Accuracy and Efficiency with Reservoir Computers:}}} While ML model-agnostic, our implementation leverages the simplicity, efficiency, and low computational costs of RCs, extending the utility of traditional RCs to more complex forecasting scenarios requiring generalization.

Unlike prior work that relies on explicit context indicators/labels to capture the climate of unseen systems\cite{Patel2021_NonstationaryRC,Kong2021_MLSystemCollapse,Koglmayr2024_ParamAwareNGRC,Panahi2024_AdaptableRC,Kong2024_IndexBasedRC}, METAFORS requires only short cue signals to construct and initialize forecasting models directly from observations. This label-free learning makes METAFORS versatile: it minimizes the need for domain knowledge and enables learning and prediction across systems governed by distinct functional forms without requiring additional contextual information -- a key distinction from generalization schemes that rely on contextual tags. 

While METAFORS is not tied to a specific ML architecture, our reservoir computing implementation extends the capabilities of RCs themselves. By mitigating brittleness, data intensity, and warm-up requirements, METAFORS aligns with recent efforts to enhance RC generalization -- a central challenge to industrial and scientific applications of RCs\cite{Yan2024_RC_Opps_and_Challenges}. Related methods have explored cold starting for RCs\cite{Grigoryeva2024_ColdStartRC} and LSTMs\cite{Kemeth2021_ColdStartLSTM}, but these approaches focus only on short-term forecasting and assume identical training and test dynamics. Similarly, other meta-learning frameworks often rely on specific architectures, such as convolutional networks for spatiotemporal data\cite{Wang2022_DyAd} or autoencoders for dimensionality reduction\cite{Canaday2021MARC,Kirchmeyer2022a_CoDA,Panahi2025_CriticalTransitions}, limiting their flexibility. METAFORS, by contrast, generalizes across diverse systems, cold starts memory-based models, and accommodates a range of architectures for forecasting and signal mapping.

Despite its strengths, our implementation of METAFORS has limitations that warrant further exploration. For example, it requires uniformly sampled sequential data, and adapting the scheme for irregularly sampled or multiscale data would expand its applicability. Additionally, while we demonstrate success in relatively low-dimensional nonlinear systems, scalability to high-dimensional and real-world datasets remains an open question. Future work integrating unsupervised learning techniques, such as autoencoders, could enhance its ability to extract meaningful low-dimensional representations and improve its scalability. Further, while METAFORS' data-driven modeling approach is an important strength, the integration of METAFORS with hybrid forecasting architectures -- where knowledge-based models are coupled with data-driven components -- could further enhance its utility. Exploring such integrations could enable METAFORS to address scenarios where partial knowledge of the system dynamics is available, combining the strengths of data-driven and knowledge-based approaches.

In conclusion, METAFORS represents a significant advance in data-driven forecasting of dynamical systems. By leveraging meta-learning to construct tailored forecasting models, it mitigates key limitations of traditional memory-based approaches, enabling accurate predictions from limited data. Its flexibility, efficiency, and generalization capabilities make it a powerful tool for tackling pressing forecasting challenges across fields such as climate science, neuroscience\cite{DeMatola2025_BrainStateForecasting}, and public health.
\section{Methods}\label{sec:Methods}
\noindent
In this section, we build on the earlier \textit{implementation overview} to more thoroughly detail our reservoir computing implementation of METAFORS. The supplementary material contains additional background information on forecasting time series with reservoir computers (\cref{subsec:RC_Pred}), and details of our experiment setups (\cref{sec:Experimental_Design}) and baseline comparison methods (\cref{sec:Benchmark_Details}).

\subsection{Our reservoir-computing implementation of METAFORS} \label{subsec:METAFORS_Method}
\noindent
METAFORS uses two levels of learning to build and cold-start tailored forecasting models for data-limited dynamical systems by leveraging a library of models trained on potentially related long time series. Here, we use reservoir computers (RCs) for both learning levels. In the first level, we construct a library of forecaster RCs by training a different output layer for the `forecaster reservoir' on each available long signal. In the second level, a `signal mapper' RC, draws on all the dynamics represented in the library of forecaster RCs, and a short cue signal to both construct and initialize a suitable forecaster RC for that cue. Our training scheme for the signal mapper RC has elements in common with schemes used in RC-based similarity learning that have been applied to image recognition and classification\cite{Krishnagopal, schaetti}. In our case, the signal mapper infers similarity between short observed time series and learns a mapping from these short series to corresponding output layers (trainable parameters) and initial reservoir states (cold-start vectors).

We believe that RCs are a particularly strong choice for building the forecaster used in the first level of learning because of their simplicity, accuracy, and efficiency in both short-term forecasting and long-term climate replication. For the signal mapper in the second level of learning, however, we chose RCs primarily for their convenience. They remain a robust and effective option, but we nonetheless anticipate that other types of ML approaches could also perform effectively and might be advantageous in certain situations. Alternative methods might offer comparable or improved performance for the forecaster, additional flexibility or interpretability for the signal mapper, or even complementary benefits that enhance the overall framework's adaptability to different scenarios.

\subsubsection{Requirements and scope}\label{sec:METAFORS_Scope}
\noindent
METAFORS requires that we have available a short observed signal or cue, $\boldsymbol{s}_{new}$ (denoted $\boldsymbol{s}_{test}$ in the case of our experiments with simple test systems), from the dynamical system we wish to predict as well as a collection, or library, $\{\boldsymbol{L}_i\},\:{i=1,...,N_L}$, of $N_L$ long signals from systems that exhibit similar dynamics to the test system. In our RC implementation of METAFORS, the data from each time series must be sequential and sampled at even intervals, $\Delta t$, and must also be of the same dimension, which we denote $N_{sys}$. We emphasize that the library and test signals need not contain full information of the system state at each time step. $N_{sys}$ is merely the number of observables we wish to predict. We do not require that all long signals have equal duration, but we assume that each is sufficiently long to train an RC well. METAFORS is most useful when the duration of the test signal, ${t_{test}=(N_{test}-1)\Delta t}$, where $N_{test}$ is the number of data points in the signal, is insufficient to train an RC directly. METAFORS' ability to cold-start the forecaster RC is most useful when $t_{test}$ is insufficient to synchronize the forecaster reservoir state to the test signal.

\subsubsection{Constructing the library}\label{sec:Library_Construction}
\noindent
We first learn an RC representation of each system in the set of long signals, using the same forecaster reservoir layer with ${N_r = N_F}$ nodes for each. We refer to any combination of the forecaster reservoir with a trained output layer, $W_{out}$, as a forecaster RC and construct a library of forecaster RCs as follows (Fig.~\ref{fig:LARC_Schematic}A).
\begin{enumerate}[label = (2\alph*), topsep=0pt, itemsep=-1ex, partopsep=1ex, parsep=1ex]
    \item Using the same reservoir layer for each of the available long series, $\boldsymbol{L}_i(t)$, we train a forecaster RC with training time series $\boldsymbol{u}(t) = \boldsymbol{L}_i(t)$ to obtain a set of corresponding output layers, $\{W_{out}^i\}$. The output layer $W_{out}^i$ constitutes the trainable parameters, $\boldsymbol{\theta}_i$, of the forecaster for library member $i$. We also store the reservoir trajectory, $\boldsymbol{r}_i(t)$, over the fitting period of each long signal.
    \label{step:long_training_step}
    \item We divide each long signal into sub-signals consisting of $N_{test}$ sequential data points. In this work, we extract all possible short signals of length $N_{test}$ after discarding a transient of length $N_{trans}$, i.e.,
    \begin{equation}\label{eq:Short_Signals}
        \boldsymbol{s}_{ij}(k\Delta t)=\boldsymbol{L}_i(j\Delta t+k\Delta t)\\ \ \forall\ \begin{cases}1\leq i\leq N_L\\j \ge N_{trans},\\0\leq k\leq N_{test}-1 \end{cases}
    \end{equation}
    where $\boldsymbol{s}_{ij}$ denotes the $j^{th}$ sub-signal extracted from long signal $\boldsymbol{L}_i$ and $N_{trans}$ is the transient time, as in Eq.~\ref{eq:Ridge_Cost}. Here $j$ also indexes the time step at the start of the short signal since we utilize all possible short signals after the transient, although we note that METAFORS can still perform strongly when short signals are subsampled from the long library signals.  
    \item For each short signal, $\boldsymbol{s}_{ij}$, we extract a corresponding initial reservoir state\begin{equation}\label{eq:Initial_States}
        \boldsymbol{r}_{ij}(0)=\boldsymbol{r}_i(j\Delta t)
    \end{equation}
    from the stored reservoir trajectories, $\{\boldsymbol{r}_i\}$. ${\boldsymbol{r}_{ij}(0)}$ is a constructed cold-start vector, $\boldsymbol{m}_{ij}$, for the signal $\boldsymbol{s}_{ij}$ (Fig.~\ref{fig:LARC_Schematic}B).
\end{enumerate}
Each library member in METAFORS comprises a long signal and its corresponding trained forecaster model. The set of triplets of short signals, associated initial reservoir states, and trained output layers ${\{(\boldsymbol{s}_{ij}, \boldsymbol{r}_{ij}(0), W_{out}^i)\}}$ forms the training data for the signal mapper.

\subsubsection{Training the signal mapper RC}\label{sec:SM_Training}
\noindent
We train the signal mapper RC, with ${N_r=N_{SM}}$ nodes in its reservoir, to map each short signal $\boldsymbol{s}_{ij}$ to the corresponding initial reservoir state and output layer pair, $(\boldsymbol{r}_{ij}(0), W_{out}^i)$, as follows (Fig.~\ref{fig:LARC_Schematic}C).
\begin{enumerate}[label = (3\alph*), topsep=0pt, itemsep=-1ex, partopsep=1ex, parsep=1ex]
    \item \label{step:SM_Drive_Train}For each short signal, $\boldsymbol{s}_{ij}$, in the library, we set the signal mapper to have initial reservoir state $\boldsymbol{r}^{SM}(0)=\boldsymbol{0}$. We then feed the short signal into the signal mapper and store the final state, ${\boldsymbol{r}^{SM}(N_{test}\Delta t)}$, of the resulting trajectory in the reservoir state-space.
    \item \label{step:Train_SM}We use ridge regression to find a linear mapping, $W_{SM}$, from each such final reservoir state, ${\boldsymbol{r}^{SM}(N_{test}\Delta t)}$, to its corresponding pair ${(\boldsymbol{r}_{ij}(0), W_{out}^i)}$:
    \begin{equation}
    W_{SM}=PR^T\left(RR^T+\alpha_{SM}N_{short}I\right)^{-1},
    \label{eq:SM_Ridge}
    \end{equation}
    where $N_{short}$ is the number of short signals in the library (the number of training pairs), and $R$ (${N_{SM}\times N_{short}}$) and $P$ ${\left((N_{sys}+1)N_F\times N_{short}\right)}$ are the horizontal concatenations of all final signal mapper states and all target pairs, ${\boldsymbol{p}_{ij}=[\boldsymbol{r}_{ij}(0), W_{flat}^i]^T}$, respectively. $W_{flat}^i$ is an $N_F N_{sys}$-dimensional vector representation of output layer $W_{out}^i$.
\end{enumerate}

\subsubsection{Making predictions}\label{sec:SM_Prediction}
\noindent
Given a test signal, $\boldsymbol{s}_{test}$, we construct a tailored forecaster model to generate predictions (Fig.~\ref{fig:LARC_Schematic}D).
\begin{enumerate}[label = (4\alph*), topsep=0pt, itemsep=-1ex, partopsep=1ex, parsep=1ex]
    \item \label{step:SM_Drive_Pred}We feed the test signal into the signal mapper as in step \ref{step:SM_Drive_Train} and apply the mapping learned in step \ref{step:Train_SM} to extract an appropriate initial reservoir state (i.e., cold start vector), $\boldsymbol{r}_{test}(0)$, and output layer, $W_{out}^{test}$ (i.e., set of model parameters).
    \item We construct a tailored forecaster RC for the test system by combining the inferred output layer, $W_{out}^{test}$, with the forecaster reservoir. We then synchronize this RC to the test signal in open-loop mode (Fig.~\ref{fig:Res_Diagram}A) starting from the initial state ${\boldsymbol{r}(0)=\boldsymbol{r}_{test}(0)}$, and close the loop (Fig.~\ref{fig:Res_Diagram}B) after time ${t_{test}=(N_{test}-1)\Delta t}$ to forecast from the end of the test signal. The reservoir inputs are then:
    \begin{equation*}
    	\boldsymbol{u}(t)=\begin{cases}
    		\boldsymbol{s}_{test}(t), & 0\leq t\leq t_{test} \\
    		\hat{\boldsymbol{u}}(t), & t>t_{test}
    	\end{cases}.
    \end{equation*}
    Thus, ${\hat{\boldsymbol{u}}(N_{test}\Delta t)}$ is the first output of the forecaster RC to be fed back as its input.  In other words, $t_{test}$ corresponds to the forecast start time $t_0$ in \cref{subsec:RC_Pred}.
\end{enumerate}
\section{References}
\bibliography{LARC_Bibliography.bib}
\begin{acknowledgments}\label{sec:Acknowledgements}
\noindent
We thank Daniel Canaday for helpful conversations, insights, and suggestions. We also acknowledge the University of Maryland supercomputing resources (\href{http://hpcc.umd.edu}{http://hpcc.umd.edu}) made available for conducting the research reported in this paper. This work was supported in part by DARPA under contract W31P4Q-20-C-0077. The contributions of D.N., of M.G., and of E.O. and B.H. were also supported, respectively, by the National Science Foundation Graduate Research Fellowship Program under Grant No. DGE 1840340, by ONR Grant No. N000142212656, and by ONR Grant No.N00014-22-1-2319. Any opinions, findings, and conclusions or recommendations expressed in this material are those of the authors and do not necessarily reflect the views of the National Science Foundation, the Department of Defense, or the U.S. Government.
\end{acknowledgments}
\section{Code Availability:}\label{sec:Code}
\noindent
Codes used to run the numerical experiments performed for this paper are available at the following public repository: \href{https://github.com/nortondeclan/METAFORS}{https://github.com/nortondeclan/METAFORS}.
\supplementarysection

\subsection{ Forecasting with reservoir computers} \label{subsec:RC_Pred}
\noindent
The specific reservoir computing architecture that we employ is constructed after the one proposed by Jaeger and Haas in 2004\cite{JaegarHaas}. They referred to their version of a reservoir computer (RC) as an `echo state network,' but the two terms have been used interchangeably in many contexts. In their seminal work, Jaeger and Haas demonstrated the use of RCs in predicting and processing time series.

The central component of an RC is a recurrent neural network, the reservoir. Each of its nodes, indexed by $i$, has an associated continuous-valued, time-dependent activation level $r_i(t)$. For a reservoir of size $N_r$ nodes, we refer to the vector ${\boldsymbol{r}(t)=[r_1(t),...,r_{N_r}(t)]^T}$, containing the activations of all of its nodes, as the reservoir state at time $t$. It evolves in response to an input signal according to a dynamical equation with a fixed discrete time step, $\Delta t$:
\begin{equation}
\label{eq:Open_Res_Update}
\begin{aligned}
    \boldsymbol{r}(t+\Delta t)= & (1-\lambda)\boldsymbol{r}(t) \\ & +\lambda\tanh(A\boldsymbol{r}(t)+B\boldsymbol{u}(t)+\boldsymbol{c}),
\end{aligned}
\end{equation}
where the $\tanh()$ function is applied element-wise, and the directed and weighted ${N_r\times N_r}$ adjacency matrix $A$ specifies the strength and sign of interactions between each pair of reservoir nodes. The $N_{in}$-dimensional input ${\boldsymbol{u}(t)}$ is coupled to the reservoir nodes at time $t$ via an ${N_r\times N_{in}}$ input weight matrix $B$. An $N_r$-dimensional random vector of biases, $\boldsymbol{c}$, serves to break symmetries in the dynamics of the reservoir nodes. We say that the reservoir has memory if ${\boldsymbol{r}(t + \Delta t)}$ depends on the reservoir past history of inputs, ${\boldsymbol{u}(t-m\Delta t)}$ for ${m>0}$, and (because, by Eq.~\ref{eq:Open_Res_Update}, ${\boldsymbol{r}(t)}$ depends on ${\boldsymbol{u}(t-\Delta t)}$, ${\boldsymbol{r}(t-\Delta t)}$ depends on ${\boldsymbol{u}(t-2\Delta t)}$, and so on) this will be the case if the right hand side of Eq.~\ref{eq:Open_Res_Update} explicitly depends on ${\boldsymbol{r}(t)}$ (i.e., if ${(1-\lambda)}$ and/or the matrix $A$ are nonzero). The leakage rate, $\lambda$, thus influences the time-scale on which the reservoir state evolves; when the leakage rate is zero, ${\boldsymbol{r}(t+\Delta t)=\boldsymbol{r}(t)\ \forall\ t}$ and the reservoir does not evolve; when the leakage rate is one, the first term in Eq.~\ref{eq:Open_Res_Update} is zero and the reservoir `forgets' its previous states more rapidly.

We use the \textit{rescompy} python package\cite{rescompy} to construct our RCs. The adjacency matrix, $A$, of each reservoir is a (sparse) random directed network with connection probability ${\langle d\rangle/N_r}$ for each pair of nodes, where $\langle d\rangle$ is the mean in-degree of the network. We assign non-zero elements of $A$ random values from a uniform distribution ${U[-1, 1]}$. Then we normalize this randomly generated matrix such that its spectral radius (eigenvalue of largest absolute value) has some desired value, $\rho$. We generate a dense input matrix, $B$, and a bias vector, $\boldsymbol{c}$, by choosing each entry from the uniform distributions ${U[-\sigma,\sigma]}$ and ${U[-\psi,\psi]}$, respectively. We refer to $\sigma$ as the input strength and to $\psi$ as the bias strength.

For each step $\boldsymbol{u}(t)$ of an input series, we can construct a corresponding reservoir output, $\boldsymbol{y}(t)$, of dimension $N_{out}$, as a linear combination of the node activations resulting from its input to the reservoir:
\begin{equation}
    \boldsymbol{y}(t)=W_{out}\boldsymbol{r}(t+\Delta t).
    \label{eq:Res_Output}
\end{equation}
$W_{out}$ is the ${N_{out}\times N_r}$ matrix that determines the linear combination. We call $W_{out}$ the reservoir's output layer and refer to the combination of a reservoir (defined by its internal parameters: the adjacency matrix, $A$, input matrix, $B$, bias vector, $\boldsymbol{c}$, and leakage rate, $\lambda$) and an output layer, $W_{out}$, as a reservoir computer (RC). When training an RC for forecasting tasks, we choose its output layer so that at every time step over some training signal of $N_{train}$ evenly-spaced data points, i.e. with duration ${(N_{train}-1)\Delta t}$, the RC's output closely matches its input at the next time step:
\begin{equation}
    \boldsymbol{u}(t+\Delta t)\approx\boldsymbol{y}(t)=W_{out}\boldsymbol{r}(t+\Delta t).
    \label{eq:Training_Targets}
\end{equation}
In this case, $N_{out}=N_{in}=N_{sys}$, where $N_{sys}$ is the number of observables we wish to predict. The internal parameters of the reservoir ($A$, $B$, $\boldsymbol{c}$, and $\lambda$) are not altered during training, or afterwards. To calculate the output layer, we drive the reservoir with the training signal in the open-loop mode (Eq.~\ref{eq:Open_Res_Update} and Fig.~\ref{fig:Res_Diagram}A) and minimize the ridge-regression cost function:
\begin{equation}\sum_{n=N_{trans}}^{N_{train}-1}\frac{\| W_{out}\boldsymbol{r}(n\Delta t)-\boldsymbol{u}(n\Delta t)\|^2}{N_{fit}}+\alpha_F\|W_{out}\|^2,
    \label{eq:Ridge_Cost}
\end{equation}
where the scalar $\alpha_F$ is a (Tikhonov\cite{Tikhonov1995}) regularization parameter which prevents over-fitting, $\|\ \|$ denotes the Euclidean ($L^2$) norm, and ${N_{fit}=N_{train}-N_{trans}-1}$ is the number of input/output pairs used for fitting. We discard the first $N_{trans}$ reservoir states and target outputs as a transient to allow the reservoir state to synchronize to the input signal before fitting over the remaining time steps. The minimization problem, Eq.~\ref{eq:Ridge_Cost}, has solution
\begin{equation}
    W_{out}=YR^T\left(RR^T+\alpha_F N_{fit}I\right)^{-1},
    \label{eq:Pred_Ridge}
\end{equation}
where $I$ is the identity matrix and $Y$ (${N_{out}\times N_{fit}}$) and $R$ (${N_r\times N_{fit}}$), with n\textsuperscript{th} columns ${\boldsymbol{u}([N_{trans}+n]\Delta t)}$ and ${\boldsymbol{r}([N_{trans}+n]\Delta t)}$, respectively, are the target and reservoir state trajectories over the fitting period.

Once we have trained a reservoir computer (RC) as above, we can use it to obtain predictions, $\hat{\boldsymbol{u}}(t)$, of $\boldsymbol{u}(t)$, for values of $t > t_0 \geq (N_{train}-1)\Delta t$, where $t_0$ is the forecast start time. In the prediction phase, the RC evolves autonomously in closed-loop mode (Fig.~\ref{fig:Res_Diagram}B) by setting its input at each time step to be its output from the previous time step in a cyclic manner. The RC itself thus forms an autonomous dynamical system which evolves according to the coupled equations
\begin{subequations}\begin{equation}
\begin{aligned}
    \boldsymbol{r}(t+\Delta t) = & (1-\lambda)\boldsymbol{r}(t) \\ & +\lambda\tanh(A\boldsymbol{r}(t)+B\hat{\boldsymbol{u}}(t)+\boldsymbol{c})
\end{aligned}\label{eq:Closed_Loop_Res}\end{equation}\begin{equation}
    \hat{\boldsymbol{u}}(t+\Delta t)=W_{out}\boldsymbol{r}(t+\Delta t)\label{eq:Closed_Loop_u}
\end{equation}\label{eq:Closed_Loop_Updates}\end{subequations}
and mimics the system of interest.

It is worthwhile to consider the separate roles of the reservoir state, ${\boldsymbol{r}(t)}$, and output layer, $W_{out}$, in the autonomous RC system, Eq.~\ref{eq:Closed_Loop_Updates}. We give a brief outline of these roles here and point to the work of Lu, Hunt, and Ott\cite{Lu2018_Attractor_Reconstruction} for a more complete explanation and for discussion of the conditions under which this description holds.

In general, the reservoir component of an RC that has been successfully trained to accomplish its time series prediction task satisfies the \textit{echo state property}\cite{Jaeger2001_ESP,Lukosevicius_2012,Cucchi2022_Hands_On_RC,Platt2021_PredictiveGS,Platt2022_RC_for_Complex_Forecasting_Review}: if the reservoir is driven in the open-loop mode (Eq.~\ref{eq:Open_Res_Update} and Fig.~\ref{fig:Res_Diagram}A) twice with the same input signal but different initial reservoir states, it will converge in both cases to the same trajectory after sufficient time. In other words, after some transient response of the reservoir has passed, its state becomes independent of its initial condition and depends only on the history of the input. One consequence of the echo state property is that any input trajectory ${\boldsymbol{u}}$ that evolves on some manifold, $M_{sys}$ will drive the reservoir to evolve along a trajectory, ${\boldsymbol{r}}$, such that once the transient period has passed,  $\boldsymbol{r}$ and $\boldsymbol{u}$ are synchronized and $\boldsymbol{r}$ lies on some corresponding manifold, $M_{res}$, in the reservoir state-space\cite{Lu2018_Attractor_Reconstruction,Platt2021_PredictiveGS,Platt2022_RC_for_Complex_Forecasting_Review}. The output layer of an RC trained on $\boldsymbol{u}$ maps points on or near $M_{res}$ to points on or near $M_{sys}$. For a fixed reservoir (defined by its internal parameters $A$, $B$, $\boldsymbol{c}$, and $\lambda$), the structures of these manifolds and, consequently, of the RC's output layer are determined solely by the dynamics of the system from which $\boldsymbol{u}$ is sampled -- the initial conditions of the reservoir and of the time series $\boldsymbol{u}$ have negligible impact if $N_{trans}$ and $N_{train}$ are sufficiently large. We may thus build the following intuition, which is central to our RC implementation of METAFORS: an output layer, $W_{out}$, of an RC encodes the dynamics of the system on the manifold $M_{sys}$; the reservoir state, ${\boldsymbol{r}(t)}$, determines its phase.

Generally, we expect the error in an RC's forecast to grow with time. There are two sources of this error:
\begin{enumerate}[label = (\arabic*), topsep=0pt, itemsep=-1ex, partopsep=1ex, parsep=1ex]
	\item Initial State Error: If the reservoir state at the start of a prediction, ${\boldsymbol{r}(t_0)}$, is too far from the manifold $M_{res}$ corresponding to the dynamics encoded by the RC's output layer, the forecast dynamics may be inconsistent with those of the output layer. If ${\boldsymbol{r}(t_0)}$ is close to $M_{res}$ but is not well synchronized to the state of the true system at time ${t=t_0}$, the forecast will be out of phase with the true system.\label{err:initial_state}
	\item Output Layer Error: Errors in the learned output layer, $W_{out}$, induce errors in the prediction ${\hat{\boldsymbol{u}}(t)= W_{out}r(t)}$.\label{err:output_layer}
\end{enumerate}
If the underlying system is chaotic, both types of errors are amplified as $t$ increases.

Typically, to initialize a forecast, we drive the reservoir in the open-loop mode (Eq.~\ref{eq:Open_Res_Update} and Fig.~\ref{fig:Res_Diagram}A) with a sync signal, $\boldsymbol{u}_{sync}$, from the true system and then switch to the closed-loop mode (Eq.~\ref{eq:Closed_Loop_Updates} and Fig.~\ref{fig:Res_Diagram}B) to forecast from the end of $\boldsymbol{u}_{sync}$. The sync signal can be significantly shorter than would be sufficient to train an RC accurately. Hence, an RC that has been trained on a long time series to accurately capture the dynamics of $\boldsymbol{u}(t)$, can be used to predict from a different initial condition by starting from a comparatively short sync signal. However, if the sync signal is too short, the initial state error \ref{err:initial_state} will lead to an inaccurate forecast even if the output layer can accurately capture the dynamics of the true system.

\subsection{ Testbeds for probing the utility of METAFORS}
\label{sec:Experimental_Design}

\subsubsection{ The logistic and Gauss iterated maps}
\label{sec:Forecaster_Training_Maps}
\noindent
We exclude periodic trajectories from the training library in our experiments with the logistic and Gauss iterated maps because these trajectories poorly sample the state space of the maps' dynamics. A non-parametric forecasting model, such as an RC, trained on an orbit with a short periodicity length may not learn a sufficiently complex representation of the dynamics. As a result, its inclusion in the training library can hamper generalization.

For each long signal in the library, we simulate the logistic/Gauss iterated map for $2000$ total iterations. We discard the first $1000$ iterations of each to ensure that the time series used for training are well-converged to the system's attractor. Each signal's remaining ${N_{train}=1000}$ points constitute a long library signal. We use the next ${N_{trans}=50}$ iterations of each signal as a transient period to synchronize the internal state of the forecaster RC to the input series, and then fit the trainable parameters of the forecaster, contained in its output layer, to the remaining $N_{fit}=950$ iterations. We forecast $N_{for}=1000$ iterations beyond the end of each test signal. Since, in these experiments, we are interested only in the long-term statistics, or climate, of a forecast, we discard the first $500$ of these predicted points before calculating cumulative probability distributions or plotting bifurcation diagrams.

\begin{table}[!h]
	\begin{ruledtabular}
		\begin{tabular}{l|l|lll|l|lll}
			&\multicolumn{4}{c|}{Forecaster}&\multicolumn{4}{c}{Signal Mapper}\\\hline
			Reservoir Size & $N_F$ & \multicolumn{3}{c|}{500} & $N_{SM}$ & \multicolumn{3}{c}{1000} \\
			Mean In-degree & $\langle d\rangle_F$ & \multicolumn{3}{c|}{3} & $\langle d\rangle_{SM}$ & \multicolumn{3}{c}{3} \\
			Spectral Radius & $\rho_F$ & 0.2 & 0.2 & 0.9 & $\rho_{SM}$ & \multicolumn{3}{c}{0.9} \\
			Input Strength & $\sigma_F$ & 2.5 & 4.0 & 0.1 & $\sigma_{SM}$ & 2.5 & 4.0 & 0.1 \\
			Bias Strength & $\psi_F$ & \multicolumn{3}{c|}{0.5} & $\psi_{SM}$ & \multicolumn{3}{c}{0.5} \\
			Leakage Rate & $\lambda_F$ & 0.2 & 0.2 & 0.1 & $\lambda_{SM}$ & \multicolumn{3}{c}{0.1} \\ 
			Regularization & $\alpha_F$ & \multicolumn{3}{c|}{$10^{-6}$} & $\alpha_{SM}$ & \multicolumn{3}{c}{$10^{-8}$}
		\end{tabular}
	\end{ruledtabular}
	\caption{\justifying \textbf{Reservoir Hyperparameters}\newline For entries with multiple values, those values pertain, from left to right, to our experiments with the logistic map only, with the logistic and Gauss iterated maps,
    and with the Lorenz-63 equations.
    }
	\label{tab:Reservoir_Parameters}
\end{table}

As shown in ~Table~\ref{tab:Reservoir_Parameters}, the reservoir hyperparameters that we use for our experiments with the logistic map only (Fig.~\ref{fig:Logistic_Map_Results}) and for our experiments with both the logistic and Gauss iterated maps (Fig.~\ref{fig:Dual_Library}) are identical except for the input strengths, $\sigma_F$ and $\sigma_{SM}$. We chose the size, mean in-degree, and bias of both networks to have values that typically allow for reasonably accurate forecasting with reservoir computers, and performed no experiment-specific tuning of these values. We chose the leakage rate and spectral radius of the signal mapper such that its memory is sufficiently long that the final reservoir state of the signal mapper, after it has received a test signal as input, is influenced by many, or all, of the data points in the test signal. We chose the input strength of both the forecaster and signal mapper networks such that the average of the product of the input weight matrix and the input data from the library is approximately one, ${\langle B\boldsymbol{u}_i(t)\rangle_{i,t}=\langle B\boldsymbol{L}_i(t)\rangle_{i,t}\approx1}$. We chose the remaining hyperparameters -- the regularization strengths, forecaster leakage rate, and forecaster spectral radius -- by coarse hand-tuning to allow for good, but not necessarily optimal performance. While more robust hyperparameter tuning may improve performance overall, our priority is to compare the relative performances of simple baseline methods on familiar systems, rather than to obtain highly optimized forecasts.

\subsubsection{ The Lorenz-63 equations}
\label{sec:Forecaster_Training_Lorenz}
\noindent
To generate the library and test signals, we integrate Eq.~\ref{eq:Lorenz}, using a fourth-order Runge-Kutta scheme with fixed time step $\Delta t=0.01$, with different values of $v_1$ and $\omega_t$ and starting from randomly chosen initial conditions. We hold the Lorenz parameters ${v_2=28}$ and ${v_3=8/3}$ fixed. To ensure that the full duration of each signal lies on its corresponding system's attractor and contains no transient behavior, we discard the first $1000$ points of each generated trajectory. Except where otherwise indicated, the long library signals consist of ${N_{train}=6000}$ sequential data-points. We use the first ${N_{trans}=1000}$ points of each signal as a transient to synchronize the internal state of the forecaster reservoir to the signal and fit the forecaster's output layer to the remaining ${N_{fit}=5000}$ points. We choose the parameters $v_1$ and $\omega_t$ of the library members randomly from the uniform distributions ${\omega_t\in U[0.75, 1.25]}$ and ${v_1\in U[7.5,12.5]}$. We choose the test systems' parameters to form a rectangular grid spanning the region defined by ${0.7\leq\omega_t\leq 1.3}$ and ${7\leq v_1\leq 13}$. The resolution of this grid is ${30\times30}$ in Fig.~\ref{fig:Lorenz_Results}B and ${25\times25}$ for the other figures. We forecast ${N_{for}=3000}$ data-points beyond the end of each test signal. The reservoir hyperparameters (Table~\ref{tab:Reservoir_Parameters}) for our experiments with the Lorenz-63 Equations were chosen as described for our experiments with the logistic map only and with both the logistic and Gauss iterated maps.

\subsection{ Baseline methods for comparison}
\label{sec:Benchmark_Details}
\noindent
Except in Fig.~\ref{fig:Lorenz_Results}(D), we use the forecaster reservoir hyperparameters given in Table~\ref{tab:Reservoir_Parameters} for METAFORS and for all of the baseline methods included in our results. We also discard the same number of data points, $N_{trans}$, before fitting the forecaster RC's trainable parameters for all methods except \textit{Training on the Test Signal}. Below we provide additional details for three baseline approaches that are not fully specified in the main text.

\subsubsection{ Training on the test signal}
\label{subsec:Direct_Training}
\noindent
We train the forecaster reservoir separately on each short test signal. The reservoir has a zero-vector initial state, ${\boldsymbol{r}(0)=\boldsymbol{0}}$, at the start of each test signal and we discard the first ${N_{trans}}$ data points before fitting. For our results with the logistic and Gauss iterated maps,
\begin{equation*}
    N_{trans}=\begin{cases}
        0 & N_{test}=2 \\
        \lfloor N_{test}/2\rfloor & N_{test} < 10\\
        5 & N_{test}\geq10
    \end{cases},
\end{equation*}
where ${\lfloor\rfloor}$ denotes floor division. For our results with the Lorenz-63 equations,
\begin{equation*}
    N_{trans}=\begin{cases}
        \lfloor N_{test}/10\rfloor & N_{test} < 100\\
        10 & N_{test}\geq100
    \end{cases},
\end{equation*}
because we use a forecaster reservoir with a longer memory than in our results with the logistic and Gauss iterated maps. Training on the test signal directly is not possible if the test signal contains only a single data point (${N_{test}=1}$).

\subsubsection{ Multi-task learning}
\label{subsec:Multitask}
\noindent
We train the forecaster reservoir on the union of all long library signals. We set the reservoir state to zero at the start of every library signal and discard the first ${N_{trans}}$ data points of each before fitting the forecaster's output layer. This ensures that the reservoir is well synchronized to each long signal for all the data points used for fitting.

\subsubsection{ Interpolated/extrapolated forecaster}
\label{subsec:Interpolation}
\noindent
This method relies on knowledge of the dynamical parameters for each of the training and test systems. In our experiments with the logistic and Gauss iterated maps (Fig.~\ref{fig:Logistic_Map_Results} and Fig.~\ref{fig:Dual_Library}), we implement this method as follows. If the dynamical parameter of the test system is within the range of the the dynamical parameters of the library models, we perform element-wise linear interpolation between the model parameters of the nearest library member with dynamical parameter greater than that of the test system and the nearest library member with dynamical parameter less than that of the test system. If the test system's dynamical parameter is beyond the range of the library members, we perform element-wise linear extrapolation using the nearest two library members.

In our experiments with the Lorenz-63 equations (Fig.~\ref{fig:Lorenz_Results}), we rescale the parameters $\omega_t$ and $v_1$ associated with each of the library members such that they span a unit interval along both axes. If the dynamical parameters of the test system are within the convex hull of the library, we triangulate the test parameters with respect to those of the library members. Then we perform linear barycentric interpolation of the corresponding forecasters’ trainable model parameters. If the test system is outside the convex hull of the library, we use the forecaster RC of the nearest library member.

%
%

%



\end{document}